\documentclass[lettersize,journal]{IEEEtran}
\usepackage[T1]{fontenc}
\usepackage{amsmath,amsfonts}
\usepackage{algorithm}
\usepackage{array}
\usepackage[caption=false,font=normalsize,labelfont=sf,textfont=sf]{subfig}
\usepackage{textcomp}
\usepackage{stfloats}
\usepackage{url}
\usepackage{verbatim}
\usepackage{graphicx}
\usepackage{cite}
\usepackage{graphicx}%
\usepackage{multirow}%
\usepackage{amsmath,amssymb,amsfonts}%
\usepackage{amsthm}%
\usepackage{mathrsfs}%
\usepackage{xcolor}%
\usepackage{textcomp}%
\usepackage{manyfoot}%
\usepackage{booktabs}%
\usepackage{algpseudocode}%
\usepackage{listings}%
\usepackage{float}
\usepackage[cal=cm]{mathalfa}
\usepackage{url}
\usepackage{colortbl}
\usepackage{xcolor}
\usepackage{threeparttable}
\usepackage{enumitem}
\begin{document}

\title{KEPLA: A Knowledge-Enhanced Deep Learning Framework for Accurate Protein-Ligand Binding Affinity Prediction}

\author{Han Liu,
        Keyan Ding,
        Peilin Chen,
        Yinwei Wei,
        Liqiang Nie,~\IEEEmembership{Senior Member,~IEEE},
        Dapeng Wu,~\IEEEmembership{Fellow,~IEEE},
        and Shiqi Wang,~\IEEEmembership{Senior Member,~IEEE}
\thanks{Han Liu, Peilin Chen, Dapeng Wu, and Shiqi Wang are with the Department of Computer Science, City University of Hong Kong, Hong Kong (SAR), China (e-mail: hanliu.sdu@gmail.com; plchen3@cityu.edu.hk; dapengwu@cityu.edu.hk; shiqwang@cityu.edu.hk).}
\thanks{Keyan Ding is with the ZJU-Hangzhou Global Scientific and Technological Innovation Center, Zhejiang University, Hangzhou, China (e-mail: dingkeyan@zju.edu.cn).}
\thanks{Yinwei Wei is with the School of Software, Shandong University, Jinan, China (e-mail: weiyinwei@hotmail.com).}
\thanks{Liqiang Nie is with the College of Informatics, Harbin Institute of Technology (Shenzhen), Shenzhen, China (e-mail: nieliqiang@gmail.com).}
\thanks{\textit{(Corresponding author: Shiqi Wang.)}}
}

\markboth{}%
{Shell \MakeLowercase{\textit{et al.}}: A Sample Article Using IEEEtran.cls for IEEE Journals}


\maketitle

\begin{abstract}
Accurate prediction of protein-ligand binding affinity is critical for drug discovery. While recent deep learning approaches have demonstrated promising results, they often rely solely on structural features of proteins and ligands, overlooking their valuable biochemical knowledge associated with binding affinity. To address this limitation, we propose KEPLA, a novel deep learning framework that explicitly integrates prior knowledge from Gene Ontology and ligand properties to enhance prediction performance. KEPLA takes protein sequences and ligand molecular graphs as input and optimizes two complementary objectives: (1) aligning global representations with knowledge graph relations to capture domain-specific biochemical insights, and (2) leveraging cross attention between local representations to construct fine-grained joint embeddings for prediction. Experiments on two benchmark datasets across both in-domain and cross-domain scenarios demonstrate that KEPLA consistently outperforms state-of-the-art baselines. Furthermore, interpretability analyses based on knowledge graph relations and cross attention maps provide valuable insights into the underlying predictive mechanisms. The code and datasets of KEPLA are available at the GitHub repository (\url{https://github.com/hanliu95/KEPLA}).
\end{abstract}


\section{Introduction}

\IEEEPARstart{T}{he} prediction of protein-ligand binding affinity (PLA) is a key task in computational drug discovery~\cite{volkov2022frustration}. Ligands, including small molecules and biologics, can bind to proteins as agonists or inhibitors. Their interaction strength is measured by binding affinity, which reflects how effectively a ligand can modulate a target protein. Although experimental assays can measure binding affinity, they are labor-intensive and time-consuming. Therefore, deep learning approaches have attracted increasing attention for rapid affinity prediction and candidate ranking in early-stage drug discovery~\cite{sadybekov2023computational,luo2017network}.

Deep learning approaches for PLA prediction can be broadly divided into interaction-based and interaction-free methods~\cite{rezaei2020deep}.
Interaction-based methods make predictions based on the known three-dimensional (3D) structures of complexes and the physical interactions between proteins and ligands~\cite{li2022dyscore,li2025knowledge,lai2024interformer}. However, their applicability is limited when 3D complex structures are unavailable for unknown protein-ligand pairs. In contrast, without relying on direct physical interactions, interaction-free methods infer binding affinity from protein and ligand lower-dimensional data, such as amino acid sequence and simplified molecular input line entry system (SMILES) string~\cite{weininger1988smiles}.
Regarding PLA prediction as a regression task, interaction-free approaches first feed the protein and ligand inputs into different encoders such as convolutional neural network (CNN)~\cite{ozturk2018deepdta,lee2019deepconv}, graph neural network (GNN)~\cite{jiang2020drug,yang2022mgraphdta}, and Transformer~\cite{vaswani2017attention,chen2020transformercpi,huang2021moltrans}. With these advanced deep learning techniques, data-driven representations can be automatically learned from large-scale protein and ligand data. Subsequently, the binding affinities between proteins and ligands are predicted through a decoder that incorporates and processes their representations. Therefore, interaction-free methods are more broadly applicable due to their lower input requirements.

Despite their promising results, existing interaction-free models still suffer from two key limitations: (1) their reliance solely on structural information creates performance bottlenecks, and (2) their predictions often lack scientific interpretability. To overcome these limitations, we argue that interaction-free models should incorporate biochemical factual knowledge. Specifically, protein-related knowledge, such as Gene Ontology (GO)~\cite{ashburner2000gene} annotations, provides mechanistic priors about molecular function, biological processes, and cellular components. These are linked to binding pockets, catalytic residues, and cofactors that are important for PLA. Ligand-related knowledge, such as ligand property (LP), captures key chemical factors influencing binding affinity, including hydrogen-bonding capacity, charge distribution, and hydrophobicity. Therefore, such biochemical knowledge can complement structural representations and provide a more interpretable basis for PLA prediction.

However, integrating such domain knowledge presents two key challenges:  (1) how to organize the diverse and complex knowledge into a form that the model can easily process, and (2) how to enable the model to sufficiently capture the organized knowledge. To address the first challenge, knowledge graphs (KGs)~\cite{ye2021unified} provide a natural way to organize biochemical knowledge. KGs constructed from GO and LP can systematically describe biochemical knowledge of proteins and ligands via the form of \textit{entity-relation-entity} triples. Correspondingly, KG embedding techniques~\cite{wang2017knowledge} can map proteins, ligands, and related entities into continuous vector spaces, preserving both semantic and relational information. 
For the second challenge, directly using KG-derived embeddings with structural representations is a straightforward solution. However, this approach falls short for knowledge-sparse entities, as KG embeddings cannot substitute structural encoders when only amino acid sequences or SMILES strings are available. Therefore, using KG supervision to enhance structural encoders is a more suitable strategy. To this end, we propose a joint learning framework that bridges structural encoding and KG embedding through multi-objective learning, ensuring that biochemical knowledge is effectively injected during the representation learning process.

In this paper, we propose a knowledge-enhanced protein-ligand binding affinity prediction (KEPLA) model, which is the first general framework to achieve deep integration of biochemical factual knowledge into the task of PLA prediction. The overall architecture of KEPLA follows an encoder-decoder paradigm. Specifically, it employs hybrid encoders---evolutionary sequence model (ESM)~\cite{lin2023evolutionary} for 1D amino acid sequences and graph convolutional networks (GCNs)~\cite{mastropietro2023learning} for 2D molecular graphs---to generate both global and local representations of proteins and ligands. To ensure the encoders effectively capture and leverage biochemical knowledge for the task, KEPLA jointly trains the encoded protein and ligand representations on two objectives: KG embedding and PLA prediction. For the KG embedding objective, global representations are directly used as the entity embeddings of our constructed KG, which covers protein-GO and ligand-LP knowledge. Then they are optimized alongside other learnable entity and relation embeddings using standard KG embedding techniques. For the PLA prediction objective, local representations are processed by a pairwise interaction module, which uses a cross attention network~\cite{hou2019cross} to derive protein-ligand joint representations. The joint representations are then decoded by a multilayer perceptron (MLP) decoder to produce the final affinity prediction. 
Extensive experiments under different settings validate the effectiveness of KEPLA against state-of-the-art methods. Moreover, its knowledge-grounded framework enhances the interpretability of PLA prediction.

To summarize, our contributions are as follows: (1) We propose KEPLA, the ﬁrst knowledge-enhanced PLA predicting model that incorporates a KG with the interaction-free framework, achieving notable performance improvements. (2) By leveraging the KG and cross attention mechanism, KEPLA provides an interpretable prediction through relevant knowledge and attention visualization, moving beyond traditional black-box outputs. (3) We construct and release a novel KG dataset based on PDBbind, promoting the research on PLA prediction.

\section{Related Work}
\subsection{Protein-Ligand Binding Affinity
Prediction}
Existing machine learning approaches for PLA prediction can be broadly categorized into two groups based on their core assumptions: interaction-free methods and interaction-based methods.

Interaction-free methods predict affinity without requiring 3D complex structures or explicit interaction information~\cite{ballester2010machine}.~\cite{ballester2010machine}. For instance, DeepDTA~\cite{ozturk2018deepdta} directly employs raw ligand and protein sequences as inputs and utilizes a dual-channel CNN for feature extraction. GraphDTA~\cite{nguyen2021graphdta} and MGraphDTA~\cite{yang2022mgraphdta} represent ligands as 2D molecular graphs to preserve topological information, while DGraphDTA~\cite{jiang2020drug} enhances 1D protein sequence representations through 2D distance maps or protein graphs. Although these models offer computational efficiency by omitting 3D and atomic-level interactions, their generalization ability and interpretability remain limited~\cite{bai2023interpretable}.

In contrast, interaction-based methods incorporate 3D structures and atomic interactions to improve prediction. Early approaches, such as OnionNet~\cite{zheng2019onionnet}, typically employ atom-atom pair features for model training, while mathematical tools like topological data analysis have also been introduced for molecular characterization~\cite{lim2019predicting,yang2019analyzing}. More recent techniques based on GNNs or 3D-CNNs simplify input representations by using fundamental atomic features such as element types and pairwise distances. For example, Pafnucy~\cite{stepniewska2018development} adopts 3D-CNNs to extract spatial features from voxelized 3D grids, though their efficiency is hindered by sparse voxel occupancy. Recently, interaction graph neural network (IGNN) models, including SIGN~\cite{li2021structure} and GIANT~\cite{li2024giant}, have attracted attention for their relational reasoning capabilities, learning intra- and inter-molecular interactions in a staged or type-specific manner. However, these models are computationally expensive and rely heavily on accurate complex structures, limiting their applicability to novel targets or incomplete interaction data.

\subsection{Knowledge Graph}
KG is a structured semantic network that reveals relations between entities~\cite{wang2017knowledge}.
In recent years, KGs have been gradually introduced into AI4S fields such as protein and molecular research as an effective means of integrating multi-source biological information. By organizing heterogeneous entities and relations---such as proteins, small molecule targets, biological pathways, and experimental data—into structured graphs, KGs provide rich background knowledge and semantic constraints for AI4S tasks~\cite{su2022biomedical}. 

In the field of PLA prediction, existing methods have attempted to jointly model prior knowledge (such as protein families, functional sites, molecular substructures, and pathway relations) with sequence and structural features~\cite{ma2022kg}. By leveraging KG embeddings or GNNs, these methods incorporate global biological context into molecular representations, thereby improving prediction performance compared to relying solely on sequences or structures. Such approaches typically combine KG representations with molecular fingerprints, protein sequence embeddings, or 3D conformational features through concatenation or attention-based fusion, finally obtaining holistic representations that integrate both structural and biological information.

However, effectively integrating symbolic KG knowledge with structural features remains challenging, especially for knowledge-sparse or novel targets. To address these issues, we propose a KG-enhanced framework that integrates semantic knowledge with structural representation learning for PLA prediction.
\section{Methods}
\begin{figure*}[!t]
\centering
\includegraphics[width=0.9\linewidth]{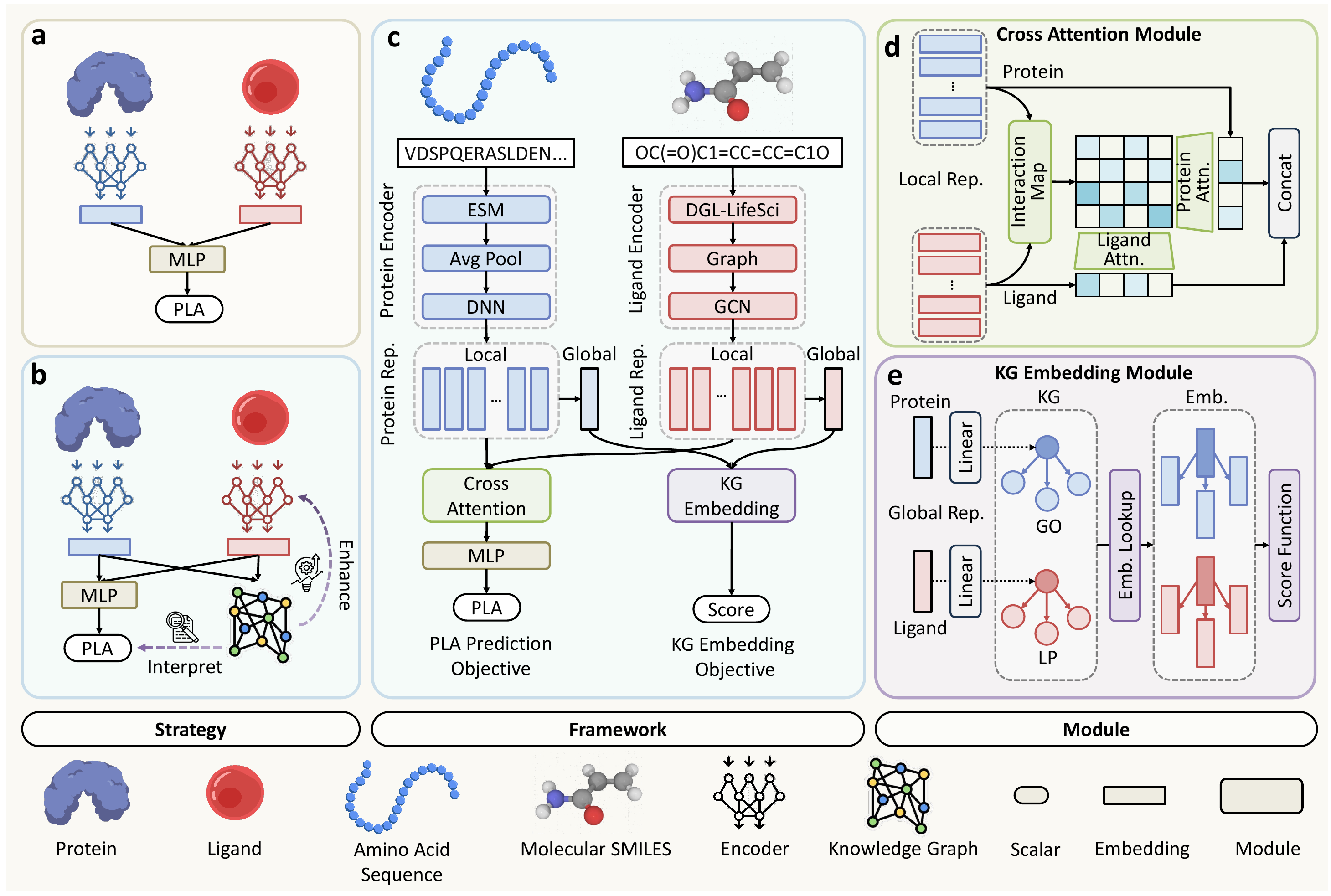}
\caption{\textbf{Overview of the KEPLA framework.} \textbf{(a)} Existing PLA prediction strategy. \textbf{(b)} KEPLA strategy, which incorporates a biochemical KG to enhance protein and ligand encoding. \textbf{(c)} KEPLA framework with two encoders: an ESM-based encoder for protein sequences and a GCN-based encoder for ligand SMILES, producing local and global representations. Global representations are used for KG embedding, while local representations are fed into the cross attention module for PLA prediction. \textbf{(d)} Cross attention module, where an interaction map derived from protein-ligand local representations enables fine-grained feature fusion, followed by an MLP decoder for affinity prediction. \textbf{(e)} KG embedding module, which projects global representations into a semantic space and evaluates triplet plausibility via a scoring function.}\label{fig_overview}
\end{figure*}

\subsection{Problem Formulation and Architecture}

In PLA prediction, each protein sequence is represented as $\mathcal{P} = (a_1, a_2, ..., a_K)$, where $a_i$ denotes one of the 23 amino acids. Each ligand is represented as a 2D molecular graph $\mathcal{G} = (\mathcal{V}, \mathcal{E})$ converted from its SMILES string~\cite{weininger1988smiles}, where $\mathcal{V}$ and $\mathcal{E}$ denote atoms and chemical bonds, respectively. The task is to learn a model that maps $\mathcal{P} \times \mathcal{G}$ to a continuous binding affinity value $y \in \mathbb{R}$. To incorporate biochemical factual knowledge, we further define the KG as $\mathcal{K}=(\mathcal{U},\mathcal{R},\mathcal{S})$, where $\mathcal{U}$ includes protein, ligand, GO, and LP entities, $\mathcal{R}$ denotes their biochemical relations, and $\mathcal{S}=\{(h,r,t)\mid h,t\in\mathcal{U}, r\in\mathcal{R}\}$ denotes the set of triples. In this work, the triples mainly include protein-GO and ligand-LP relations. Thus, the knowledge-enhanced PLA task is formulated as learning $f(\mathcal{P},\mathcal{G}, \mathcal{K}) \rightarrow y$.
Based on this formulation, the proposed KEPLA framework is illustrated in Figure~\ref{fig_overview}. Subsequently, we present a detailed elucidation of our method.


\subsection{Protein Sequence Encoding}
The protein encoder is built upon the pre-trained ESM model~\cite{lin2023evolutionary}, augmented with a multi-layer neural network. It transforms an input amino acid sequence into a set of latent vectors, each representing a local fragment of the protein. ESM, a widely recognized protein language model, has demonstrated strong performance in predicting protein structure, function, and other properties from individual sequences. To fully leverage ESM's powerful encoding capabilities, we use it as the backbone of our protein encoder. Given a target protein sequence $\mathcal{P}$, we input it into the pre-trained ESM model and extract the final hidden state as the initial feature matrix:
\begin{equation}
    \mathbf{M}_p = \text{ESM}(\mathcal{P}),
\end{equation}
where $\mathbf{M}_p \in \mathbb{R}^{D_p \times K}$, with $D_p$ denoting the output dimension of ESM and $K$ representing the maximum allowed sequence length. To support batch training, protein sequences longer than $K$ are truncated, while shorter ones are zero-padded.

Since the initial features are high-dimensional and task-agnostic, they require further processing. To compress the representation and retain local sequence information, we apply average pooling over non-overlapping fragments of size $s$, yielding $s$-mer residue-level features:
\begin{equation}
    \mathbf{X}_p = \text{AvgPool}_s(\mathbf{M}_p),
\end{equation}
where $\mathbf{X}_p\in\mathbb{R}^{D_p \times M}$, with $M = K/s$ denoting the number of local fragments. Next, a multi-layer deep neural network (DNN) is employed to adapt these features to our task and extract higher-level abstractions of local protein fragments. The encoding process is defined as:
\begin{equation}
    \mathbf{H}_p^{(l+1)} = \sigma(\text{DNN}(\mathbf{W}_n^{(l)}, \mathbf{b}_n^{(l)}, \mathbf{H}_p^{(l)})), 
\end{equation}
where $\mathbf{W}_n^{(l)}$ and $\mathbf{b}_n^{(l)}$ are the learnable weights and biases of the $l$-th DNN layer, $\mathbf{H}_p^{(l)}$ is the $l$-th hidden protein representation, and $\sigma(\cdot)$ is a non-linear activation function (ReLU in our case).  The initial input is set as $\mathbf{H}_p^{(0)} = \mathbf{X}_p$.
The final output, $\mathbf{H}_p \in \mathbb{R}^{D \times M}$, is adopted as the protein local representation, where $D$ is the representation dimension. In a simple yet effective manner, the protein global representation is then computed as $\text{Average}(\mathbf{H}_p) \in \mathbb{R}^D$, where $\text{Average}(\cdot)$ denotes the averaging operation over all local representations.

\subsection{Molecular Graph Encoding}
For each ligand compound, we convert its SMILES string into a 2D molecular graph $\mathcal{G}$. To represent the node (atom) information in $\mathcal{G}$, we initialize each atom using its chemical properties, following the implementation in the DGL-LifeSci library~\cite{li2021dgl}. Each atom is initially represented as a 74-dimensional integer vector describing eight pieces of information: atom type, degree, number of implicit hydrogens, formal charge, number of radical electrons, hybridization state, number of total hydrogen atoms, and aromaticity. This initialization is formally written as:
\begin{equation}
    \mathbf{M}_d = \text{DGL-LifeSci}(\mathcal{G}),
\end{equation}
resulting in the initial node feature matrix $\mathbf{M}_d \in \mathbb{R}^{74 \times N}$, where $N$ is the maximum allowed number of atoms per molecule. Molecules with fewer atoms are padded with virtual nodes filled with zeros to ensure consistent dimensions for batch training. Next, a linear transformation is applied to project the integer feature matrix into a real-valued one: 
\begin{equation}
    \mathbf{X}_d=\mathbf{W}\mathbf{M}_d,
\end{equation}
producing a dense feature matrix $\mathbf{X}_d \in \mathbb{R}^{D_d \times N}$, with $D_d$ denoting the feature dimension.

To learn expressive representations of ligand molecules, we employ a multi-layer GCN block. GCN extends convolution operations to irregular graph structures by updating each node’s representation through aggregation of its neighboring nodes, effectively capturing molecular substructure information. We retain node-level representations to enable subsequent fine-grained modeling of interactions with protein fragments. The GCN-based ligand encoder is formulated as:
\begin{equation}
    \mathbf{H}_d^{(l+1)}=\sigma(\text{GCN}(\mathbf{A}, \mathbf{W}_g^{(l)}, \mathbf{b}_g^{(l)}, \mathbf{H}_d^{(l)})),
\end{equation}
where $\mathbf{W}_g^{(l)}$ and $\mathbf{b}_g^{(l)}$ are the learnable weight matrix and bias vector of the $l$-th GCN layer, $\mathbf{H}_d^{(l)}$ is the $l$-th hidden node representation, $\mathbf{A}$ is the adjacency matrix of the graph $\mathcal{G}$ (augmented with self-loops), and $\sigma(\cdot)$ denotes a non-linear activation function (ReLU in our case). The input to the first layer is set as $\mathbf{H}_d^{(0)}=\mathbf{X}_d$.
The final output, $\mathbf{H}_d \in \mathbb{R}^{D \times N}$, is adopted as the ligand local representation. Similarly, the ligand global representation is denoted by $\text{Average}(\mathbf{H}_d) \in \mathbb{R}^D$.

\subsection{KG Embedding Objective}
Thus far, the two encoders described above only capture the 1D and 2D structural information of proteins and ligands, respectively. To further enhance these encoders with biochemical factual knowledge, we integrate their outputs into a KG embedding framework. Specifically, we treat the encoded protein and ligand representations as entity embeddings within our constructed KG and incorporate them into a joint learning process alongside other entities and relations. The remainder of this subsection details our implementation.

Our KG structurally consists of \textit{entity-relation-entity} triplets to describe relational knowledge. We define a triplet as $(h, r, t)$, where $h$ and $t$ are head and tail entities, and $r$ is the relation linking them, whose type is usually predefined by the schema.
This KG is composed of two distinct subgraphs: one centered around proteins and the other around ligands. 
\begin{itemize}[leftmargin=*]
\item \textbf{Protein KG (GO-based)}: This subgraph incorporates two types of entities: protein entities $e_{protein}$ involved in the affinity prediction task, and Gene Ontology (GO) entities $e_{GO}$ describing functional annotations. The GO entities cover three categories: Molecular Function (MF), Cellular Component (CC), and Biological Process (BP). Therefore, the protein-related triplets fall into three groups: \textit{protein-MF}, \textit{protein-CC}, and \textit{protein-BP}. 
\item \textbf{Ligand KG (LP-based)}: Similarly, this subgraph includes ligand entities $e_{ligand}$ and ligand property (LP) entities $e_{LP}$. The LP entities are further divided into Molecular Descriptors (MD), i.e., numerical attributes derived from molecular structures (e.g., counts of hydrogen bond donors/acceptors), and Chemical Features (CF), which denote qualitative attributes (e.g., hydrophobe, positively charged). Accordingly, the ligand-related triplets fall into two groups: \textit{ligand-MD} and \textit{ligand-CF}. 
\end{itemize}

Next, we involve the protein and ligand global representations in the KG embedding objective to enhance their encoding with external knowledge:
\begin{equation}
    \mathbf{h}_p = \mathbf{W}_k^1\text{Average}(\mathbf{H}_p) + \mathbf{b}_k^1,~~~\mathbf{h}_d = \mathbf{W}_k^2\text{Average}(\mathbf{H}_d) + \mathbf{b}_k^2,
\end{equation}
where $\mathbf{W}_k^*$ and $\mathbf{b}_k^*$ are learnable weight matrices and bias vectors that map the global representations into the KG embedding space. The resulting $\mathbf{h}_p$ and $\mathbf{h}_d \in \mathbb{R}^D$ serve as the semantic embeddings of the protein and ligand entities in the KG. Similar to~\cite{bordes2013translating}, the embedding sets of other entities and relations are denoted by learnable $\mathbf{T}\in\mathbb{R}^{D\times T}$ and $\mathbf{R}\in\mathbb{R}^{D\times R}$, where $T$ is the number of non-protein and non-ligand entities, and $R$ is the number of relation types.

To achieve the KG embedding objective, we train the KG by optimizing the corresponding embeddings to fit the triplet relations. The objective function is defined as:
\begin{equation}
    \mathcal{L}_{KGE}=\frac{1}{|\mathcal{S}|}\sum_{(h,r,t)\in\mathcal{S}} F(h, r, t),
\end{equation}
where $\mathcal{S}$ is the set of all KG triplets, and $(h, r, t)$ is one of the triplets. $F(\cdot)$ is the score function of the KG embedding model. 
For modeling protein and ligand subgraphs, we adopt two widely used KG embedding models.

For the protein-GO subgraph, we adopt RotatE~\cite{sunrotate} to model hierarchical and semantically heterogeneous protein-GO relations. For the ligand-LP subgraph, we use TransE~\cite{bordes2013translating}, since ligand-property relations are relatively direct and attribute-like. Preliminary testing also shows that this combination achieves stable performance while keeping the auxiliary KG module simple. The corresponding scoring functions are defined as: 
\begin{equation}
F_p(h, r, t) = ||\mathbf{h} \circ \mathbf{r} - \mathbf{t}||,
\end{equation}
\begin{equation}
F_d(h, r, t) = ||\mathbf{h} + \mathbf{r} - \mathbf{t}||,
\end{equation}
where $F_p$ and $F_d$ denote the scoring functions for the protein-GO and ligand-LP subgraphs, respectively. The operator $\circ$ denotes the Hadamard product. In both functions, $\mathbf{h}$ is the embedding of the head entity (i.e., a protein or ligand), $\mathbf{t}$ is the embedding of the tail entity (i.e., a GO or LP entity), and $\mathbf{r}$ denotes the relation embedding.

\subsection{PLA Prediction Objective}
In parallel with KG embedding, we employ a cross attention module to capture fine-grained, pairwise local interactions between protein and ligand substructures. This module comprises two main steps: (1) constructing an interaction map to measure substructure-level similarity, and (2) applying cross attention to derive a joint protein-ligand representation.

For the first step, ESM and GCN encoders have already generated local representations for protein and ligand: $\mathbf{H}_p=\{\mathbf{h}_p^1, \mathbf{h}_p^2, ..., \mathbf{h}_p^M\}$ and $\mathbf{H}_d=\{\mathbf{h}_d^1, \mathbf{h}_d^2, ..., \mathbf{h}_d^N\}$, where $M$ and $N$ are the number of fragments in a protein and the number of atoms in a ligand, respectively. Using these local representations, we construct a pairwise interaction matrix $\mathbf{V} \in \mathbb{R}^{M \times N}$:
\begin{equation}
    \mathbf{V} = \mathbf{H}_p^{\top}\mathbf{H}_d.
\end{equation}
Each element in $\mathbf{V}$ reflects the interaction strength between a specific protein fragment and a ligand atom, thus identifying potential binding regions.

To extract a joint representation informed by these local interactions, we treat the interaction matrix as a similarity matrix and compute cross attention weights over both protein and ligand substructures. Protein attention $\boldsymbol{\alpha}_{p}\in \mathbb{R}^M$ is calculated by summing the interactions across all ligand atoms and normalizing via a Softmax function:
\begin{equation}
    \boldsymbol{\alpha}_{p} = \text{Softmax}(\frac{1}{\sqrt{D}}\cdot\sigma(\frac{1}{N}\cdot\text{Sum}(\mathbf{V}))),
\end{equation}
where $\text{Sum}(\cdot)$ denotes column-wise summation in matrix $\mathbf{V}$, and the result accumulates each fragment's interactions across ligand atoms. $\sigma(\cdot)$ selects the Tanh function, and the $\text{Softmax}(\cdot)$ normalizes the attention scores into a probability distribution. The scaling factor $\sqrt{D}$ helps stabilize gradients. Ligand attention $\boldsymbol{\alpha}_{d}\in \mathbb{R}^N$ is computed analogously by transposing the interaction matrix and aggregating over protein fragments:
\begin{equation}
    \boldsymbol{\alpha}_{d} = \text{Softmax}(\frac{1}{\sqrt{D}}\cdot\sigma(\frac{1}{M}\cdot\text{Sum}(\mathbf{V^\top}))).
\end{equation}
These attention weights can provide interpretability by highlighting which protein and ligand substructures contribute most to the interaction.

Using the attention weights, we compute the weighted sum of protein and ligand local representations and concatenate them to form a joint representation:
\begin{equation}
    \mathbf{f}_{p,d} = \mathbf{H}_p\cdot\boldsymbol{\alpha}_{p} \oplus \mathbf{H}_d\cdot\boldsymbol{\alpha}_{d},
\end{equation}
where $\mathbf{f}_{p,d} \in \mathbb{R}^{2D}$ is the joint protein-ligand representation, and $\oplus$ denotes vector concatenation. We feed this joint representation into an MLP decoder to predict the binding affinity:
\begin{equation}
    \hat{y}_{p,d} = \text{MLP}(\mathbf{f}_{p,d}),
\end{equation}
where each single layer consists of its own learnable weight matrix and bias vector, followed by a ReLU activation function. The PLA objective is defined as the mean absolute error between the predicted and true binding affinities:
\begin{equation}
    \mathcal{L}_{PLA} = \frac{1}{|\mathcal{O}|}\sum_{(p,d)\in\mathcal{O}}|y_{p,d}-\hat{y}_{p,d}|,
\end{equation}
where $\mathcal{O}$ is the set of protein-ligand training pairs, and $y_{p,d}$ is the ground truth affinity.

To jointly learn structural and semantic knowledge, we combine the affinity prediction objective with the KG embedding objective. The overall training loss function to minimize is as follows:
\begin{equation}
    \mathcal{L} = \mathcal{L}_{PLA} + \beta\mathcal{L}_{KGE} + \lambda||\boldsymbol{\Theta}||_2^2,
\end{equation}
where $\boldsymbol{\Theta}$ denotes the set of all learnable parameters, $\beta$ is a trade-off hyperparameter between tasks, and $\lambda$ is the $L_2$ regularization coefficient. We use mini-batch Adam optimization~\cite{DBLP:journals/corr/KingmaB14} to train the model. For each mini-batch of sampled pairs $(p, d) \in \mathcal{O}$, we select the associated KG triplets, construct their representations, and update all model parameters via backpropagation using the gradient of the total loss.



\section{Experimental Setup}

\subsection{Datasets}
We evaluate KEPLA and baselines on publicly available PLA benchmark datasets. We mainly use PDBbind v2016~\cite{liu2015pdb}, which consists of three overlapping subsets: The \textit{general set}, containing all 13,283 protein-ligand complexes; The \textit{refined set}, comprising 4,057 higher-quality complexes selected from the general set; The \textit{core set}, including 290 top-quality complexes curated through a rigorous selection process, serving as a benchmark for testing. For convenience, we refer to the 3,767 complexes that belong to the \textit{refined set} but are not in the \textit{core set} as the \textit{refined set} in this paper. To further assess generalization ability, we also evaluate on the CSAR-HiQ~\cite{smith2011csar} dataset, an external benchmark with two subsets containing 176 and 167 complexes, sourced independently from PDBbind.

Table~\ref{tab_kg} summarizes the entity and triplet types in our constructed KG, which involves proteins and ligands from the PDBbind \textit{refined set}.
For the Protein-GO subgraph, we retrieve GO terms from the UniProt\footnote{https://www.uniprot.org} using protein UniProt IDs. The terms cover Biological Process, Cellular Component, and Molecular Function categories. We retain the top 1,000 most frequent GO terms as GO entities. Protein-GO triples are then constructed by linking each protein to its annotated GO terms. For the Ligand-LP subgraph, LP entities are extracted from RDKit-parsed ligand MOL files, including molecular descriptors and chemical feature types. Ligand-LP triples are constructed by linking each ligand to its corresponding LP entities. For proteins or ligands without available annotations, we do not construct the corresponding KG triples, while still retaining them in PLA training through their sequence or molecular graph representations.


\begin{table}[t]
\centering
\setlength{\tabcolsep}{5pt}
\caption{\textbf{Statistics of entities and triplets in the knowledge graph.}}\label{tab_kg}
\begin{tabular}{lclc}
\toprule%
Entity types & Count & Triplet types & Count \\
\midrule
Protein & 2,542 & Protein--GO & 11,859 \\
Biological Process  & 580 & Protein--Biological Process & 4,827 \\
Cellular Component & 165  & Protein--Cellular Component & 3,598\\
Molecular Function & 255 & Protein--Molecular Function & 3,434\\
Ligand & 3,181 & Ligand--LP & 48,356\\
Molecular Descriptor & 175 & Ligand--Molecular Descriptor & 28,629\\
Chemical Feature & 23 & Ligand--Chemical Feature & 19,727\\
\bottomrule
\end{tabular}
\end{table}

\begin{table*}[t]
\setlength\tabcolsep{3pt}
\centering
\caption{\textbf{In-domain performance comparison on the PDBbind and CSAR-HiQ datasets under random split.}}\label{tab_indomain}
\begin{threeparttable}
\begin{tabular}{cl|c|c|c|c|c|c|c|c}
\toprule%
\multicolumn{2}{@{}c@{}}{\multirow{2}*{Method}} & \multicolumn{4}{@{}c@{}}{PDBbind core set} & \multicolumn{4}{@{}c@{}}{CSAR-HiQ set}\\
\cmidrule{3-10}%
& & RMSE~$\downarrow$ & MAE~$\downarrow$ & SD~$\downarrow$ & R~$\uparrow$ & RMSE~$\downarrow$ & MAE~$\downarrow$ & SD~$\downarrow$ & R~$\uparrow$\\
\midrule
\multirow{6}*{\shortstack{Interaction-free\\methods}} 
& RF-Score & 1.446 (0.008) & 1.161 (0.007) & 1.335 (0.010) & 0.789 (0.003) & 1.947 (0.012) & 1.466 (0.009) & 1.796 (0.020) & 0.723 (0.007)\\
& DeepDTA & 1.639 (0.026) & 1.315 (0.023) & 1.689 (0.025) & 0.718 (0.014) & 1.983 (0.036) & 1.510 (0.059) & 2.037 (0.051) & 0.633 (0.036)\\
& GCN & 1.735 (0.034) & 1.343 (0.037) & 1.719 (0.027) & 0.613 (0.016) & 2.324 (0.079) & 1.732 (0.065) & 2.302 (0.061) & 0.464 (0.047)\\
& GAT & 1.765 (0.026) & 1.354 (0.033) & 1.740 (0.027) & 0.601 (0.016) & 2.213 (0.053) & 1.651 (0.061) & 2.215 (0.050) & 0.524 (0.032)\\
& GIN & 1.640 (0.044) & 1.261 (0.044) & 1.621 (0.036) & 0.667 (0.018) & 2.158 (0.074) & 1.624 (0.058) & 2.156 (0.088) & 0.558 (0.047)\\
& GAT-GCN & 1.562 (0.022) & 1.191 (0.016) & 1.558 (0.018) & 0.697 (0.008) & 1.980 (0.055) & 1.493 (0.046) & 1.969 (0.057) & 0.653 (0.026)\\
& DrugBAN & 1.378 (0.025) & 1.102 (0.019) & 1.328 (0.020) & 0.788 (0.015) & 1.672 (0.034) & 1.334 (0.042) & 1.728 (0.059) & 0.767 (0.039)\\
\midrule
\multirow{13}*{\shortstack{Interaction-based\\methods}} 
& Pafnucy & 1.585 (0.013) & 1.284 (0.021) & 1.563 (0.022) & 0.695 (0.011) & 1.939 (0.103) & 1.562 (0.094) & 1.885 (0.071) & 0.686 (0.027)\\
& OnionNet & 1.407 (0.034) &  1.078 (0.028) & 1.391 (0.038) & 0.768  (0.014) & 1.927 (0.071) & 1.471 (0.031) & 1.877 (0.097) & 0.690 (0.040)\\
& SGCN & 1.583 (0.033) & 1.250 (0.036) & 1.582 (0.320) & 0.686 (0.015) & 1.902 (0.063) & 1.472 (0.067) & 1.891 (0.077) & 0.686 (0.030)\\
& GraphTrans & 1.539 (0.044) & 1.182 (0.046) & 1.521 (0.042) & 0.714 (0.019) & 1.950 (0.072) & 1.508 (0.069) & 1.886 (0.083) & 0.687 (0.033)\\
& NL-GCN & 1.516 (0.019) & 1.198 (0.013) & 1.511 (0.024) & 0.720 (0.010) & 1.840 (0.024) & 1.393 (0.016) & 1.817 (0.028) & 0.716 (0.011)\\
& GNN-DTI & 1.492 (0.025) & 1.192 (0.032) & 1.471 (0.051) & 0.736 (0.021) & 1.972 (0.061) & 1.547 (0.058) & 1.834 (0.090) & 0.709 (0.035)\\
& DMPNN & 1.493 (0.016) & 1.188 (0.009) & 1.489 (0.014) & 0.729 (0.006) & 1.886 (0.026) & 1.488 (0.054) & 1.865 (0.035) & 0.697 (0.013)\\
& MAT & 1.457 (0.037) & 1.154 (0.037) & 1.445 (0.033) & 0.747 (0.013) & 1.879 (0.065) & 1.435 (0.058) & 1.816 (0.083) & 0.715 (0.030)\\
& DimeNet & 1.453 (0.027) & 1.138 (0.026) & 1.434 (0.023) & 0.752 (0.010) & 1.805 (0.036) & 1.338 (0.026) & 1.798 (0.027) & 0.723 (0.010)\\
& CMPNN & 1.408 (0.028) & 1.117 (0.031) & 1.399 (0.025) & 0.765 (0.009) & 1.839 (0.096) & 1.411 (0.064) & 1.767 (0.103) & 0.730 (0.052)\\
& SIGN & 1.316 (0.031) & 1.027 (0.025) & 1.312 (0.035) & 0.797 (0.012) & 1.735 (0.031) & 1.327 (0.040) & 1.709 (0.044) & 0.754 (0.014)\\
& GIANT & \underline{1.269 (0.020)} & 0.999 (0.018) & 1.265 (0.024) & 0.814 (0.008) & 1.666 (0.024) & 1.242 (0.030) & 1.633 (0.034) & 0.779 (0.011)\\
& KSM & 1.272 (0.018) & \underline{0.973 (0.012)} & \underline{1.260 (0.014)} & \underline{0.816 (0.004)} & \underline{1.526 (0.026)} & \underline{1.157 (0.028)} & \underline{1.562 (0.031)} & \underline{0.803 (0.010)}\\
\midrule
\rowcolor{gray!15}
Ours & KEPLA & \pmb{1.202 (0.017)} & \pmb{0.936 (0.013)} & \pmb{1.195 (0.016)} & \pmb{0.832 (0.005)} & \pmb{1.459 (0.019)} & \pmb{1.132 (0.026)} & \pmb{1.517 (0.032)} & \pmb{0.813 (0.012)}\\
\bottomrule
\end{tabular}
\end{threeparttable}
\end{table*}

\subsection{Baselines}
We compare KEPLA with a wide range of baseline methods. All baselines are implemented using official codebases. 


\begin{itemize}[leftmargin=*]
\item \textbf{RF-Score}~\cite{ballester2010machine} is a classical machine learning method that uses handcrafted intermolecular features and a random forest for affinity prediction.

\item \textbf{DeepDTA}~\cite{ozturk2018deepdta} employs CNNs to extract features from protein sequences and ligand SMILES strings, without relying on structural information.

\item \textbf{GraphDTA}~\cite{nguyen2021graphdta} models ligands as molecular graphs and proteins as sequences. It includes four variants based on different GNN backbones: GCN, GAT, GIN, and GAT-GCN.

\item \textbf{DrugBAN}~\cite{bai2023interpretable} encodes drugs and proteins using GCN and CNN, respectively, and applies bilinear attention to model their interactions without explicit 3D structure modeling.
\end{itemize}


\begin{itemize}[leftmargin=*]
\item \textbf{Pafnucy}~\cite{stepniewska2018development} is a 3D CNN-based model that learns spatial features from voxelized protein–ligand complexes.

\item \textbf{OnionNet}~\cite{zheng2019onionnet} extracts element-pair contact features organized in concentric shells and applies CNNs for affinity prediction.

\item \textbf{SGCN}~\cite{danel2020spatial} incorporates atomic coordinates into graph convolutions to explicitly model 3D molecular structures.

\item \textbf{GraphTrans}~\cite{wu2021representing} combines GNNs with Transformer-based self-attention to capture global topological dependencies.

\item \textbf{NL-GCN}~\cite{liu2021non} introduces non-local message passing with attention-based aggregation for global structure modeling.

\item \textbf{GNN-DTI}~\cite{lim2019predicting} is a geometry-aware graph attention model that incorporates interatomic distance information.

\item \textbf{DMPNN}~\cite{yang2019analyzing} performs message passing over directed bonds, emphasizing edge-level information.

\item \textbf{MAT}~\cite{maziarka2020molecule} adopts a Transformer architecture augmented with interatomic distances for molecular representation learning.

\item \textbf{DimeNet}~\cite{gasteiger_dimenet_2020} explicitly encodes angular and radial information for precise geometric modeling.

\item \textbf{CMPNN}~\cite{song2020communicative} enhances DMPNN by enabling bidirectional communication between nodes and edges.

\item \textbf{SIGN}~\cite{li2021structure} is a simplified geometric interaction network that incorporates distances and angles for affinity prediction.

\item \textbf{GIANT}~\cite{li2024giant} is a state-of-the-art PLA model that jointly models 3D geometric graphs and interaction networks.

\item \textbf{KSM}~\cite{li2025knowledge} is a state-of-the-art structure-based PLA baseline that models distance and angle information in protein-ligand complexes. It further adopts attentive pooling to refine structural patterns for binding affinity prediction.
\end{itemize}

\subsection{Evaluation Strategies and Metrics}
We evaluate prediction performance on the two public datasets using two distinct data split strategies to simulate in-domain and cross-domain scenarios. For in-domain evaluation, following~\cite{ballester2010machine}, we use the \textit{refined set} of PDBbind as the primary training data, as there is substantial overlap between the full \textit{general set} and CSAR-HiQ dataset. We randomly split the protein-ligand complexes in the \textit{refined set} into training and validation sets at a 9:1 ratio. For testing, we use the PDBbind \textit{core set} and the CSAR-HiQ dataset after removing any overlapping complexes from the \textit{refined set}.

For cross-domain evaluation, we adopt a clustering-based pair split strategy on the large-scale PDBbind dataset to simulate realistic domain shift. Proteins and ligands are clustered independently using single-linkage hierarchical clustering, ensuring a minimum inter-cluster distance $\gamma$ to avoid overly similar samples. Proteins are represented by PSC (pseudo-amino acid composition), while ligands are encoded using ECFP4 (extended connectivity fingerprint, up to four bonds). Cosine distance and Jaccard distance are used for proteins and ligands, with thresholds $\gamma = 0.001$ and $\gamma = 0.5$, respectively. We randomly select 60\% of protein clusters and 60\% of ligand clusters to form the source domain, and use the remaining clusters to construct the target domain. The source and target domains are disjoint and follow different distributions. For training and validation, we use all source data and 80\% of the target data, while the remaining 20\% of the target data is reserved for testing. This cross-domain setting is more challenging than random splits and better reflects real-world generalization in drug discovery.

To comprehensively evaluate the model performance, following~\cite{stepniewska2018development,zheng2019onionnet}, we adopt Root Mean Square Error (RMSE), Mean Absolute Error (MAE), Pearson’s correlation coefficient (R), and the standard deviation (SD) in regression to measure the prediction error. 

\subsection{Implementation}
KEPLA is implemented using Python 3.10 and PyTorch 2.1.2, along with functionalities from several auxiliary libraries, including DGL 2.4.0\footnote{https://www.dgl.ai}, DGL-LifeSci 0.3.2\footnote{https://github.com/awslabs/dgl-lifesci}, and RDKit 2024.03.5\footnote{https://www.rdkit.org}. The protein encoder includes a pre-trained ESM2\_t36\_3B model, followed by two fully connected layers with dimensions [2560, 512, 128]. The average pooling window size is set to 9. The ligand encoder consists of three GCN layers with hidden dimensions of [128, 128, 128]. We cap the maximum allowed protein sequence length at 1,080 and the maximum number of ligand atoms at 290. In the cross attention module, we use a single attention head for simplicity. The latent embedding size is set to 128, and the fully connected decoder contains 512 hidden units. The trade-off value in the loss function is set to 0.1. We set the batch size to 64 and optimize the model using the Adam optimizer with a learning rate of 1e-4. All models are trained for a maximum of 200 epochs, and the checkpoint with the lowest validation RMSE is selected for final evaluation on the test set.

\section{Experimental Results}

\subsection{In-Domain Performance Comparison}

Table~\ref{tab_indomain} summarizes the in-domain performance comparison under the random split setting. We report the mean and standard deviation of four evaluation metrics across five independent runs. Despite being an interaction-free method, KEPLA achieves the best performance on both datasets, surpassing all evaluated baselines. Specifically, KEPLA reduces RMSE by 5.50\% on the PDBbind dataset and by 4.39\% on the CSAR-HiQ dataset compared to the best-performing baseline KSM. These results demonstrate the strong predictive capability of KEPLA.

To further clarify the trade-off between accuracy and applicability, we compare two categories of baselines. Interaction-free methods are broadly applicable because they do not require co-crystallized complex structures or docked poses, but they generally lack explicit protein-ligand interaction information. In contrast, interaction-aware methods exploit complex structures or interaction features and therefore show stronger predictive performance in our experiments. Excluding KEPLA, the strongest interaction-aware baseline outperforms the strongest interaction-free baseline, confirming the benefit of explicit interaction modeling.

KEPLA follows the interaction-free setting and requires only ligand molecular graphs and protein sequences as input. Nevertheless, it achieves better performance than both existing interaction-free baselines and the evaluated interaction-aware methods, indicating that KEPLA improves applicability without sacrificing predictive accuracy. This favorable balance mainly benefits from the proposed knowledge-enhanced representation learning and local interaction modeling.

\begin{figure*}[!t]
\centering
\includegraphics[width=0.95\linewidth]{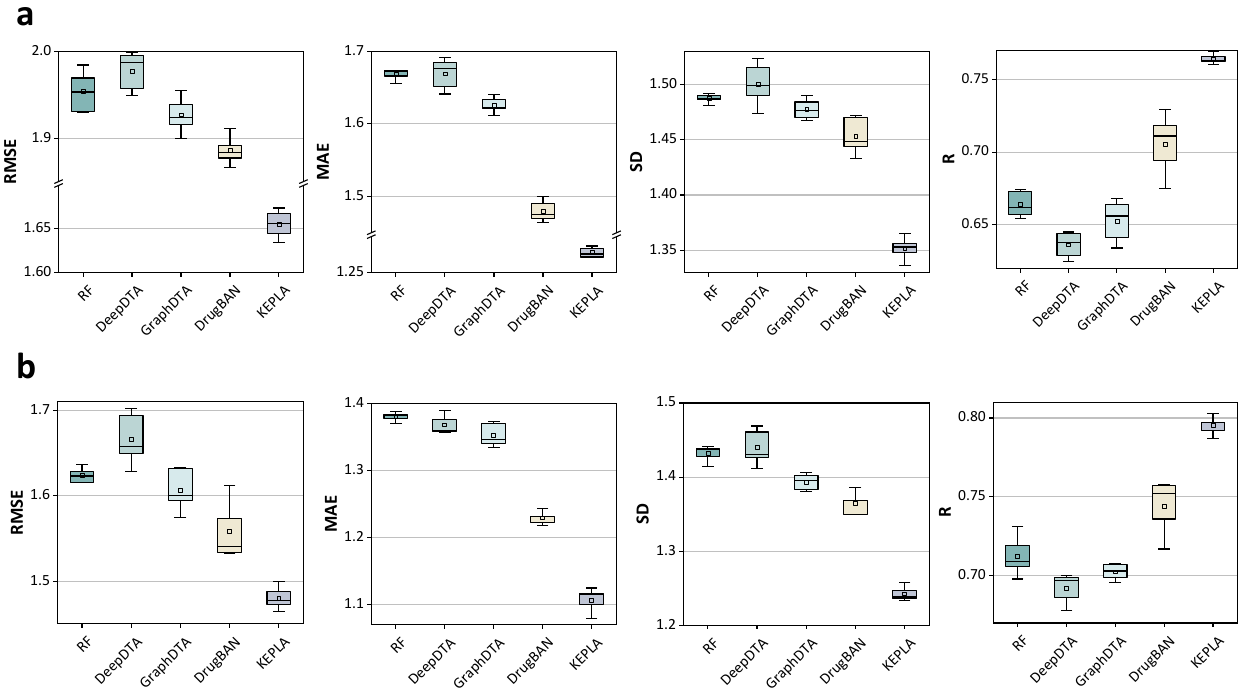}
\caption{\textbf{Cross-domain performance comparison on the PDBbind dataset (statistics over five independent runs).}~\textbf{(a)} Performance comparison of interaction-free methods under clustering-based pair split.~\textbf{(b)} Performance comparison of interaction-free methods under cold pair split. The box plots display the median as the center line and the mean as a square marker. The minimum and lower percentile indicate the worst and second-worst scores, while the maximum and upper percentile represent the best and second-best scores, respectively.}\label{fig_crossdomain}
\end{figure*}

\subsection{Cross-Domain Performance Comparison}

Beyond the random split, we further evaluate KEPLA under more challenging cross-domain settings. As a complementary evaluation, we also adopt a cold pair split, in which 70\% of proteins and ligands are randomly selected to form the training set, and the remaining 30\% are split into validation and test sets with a 3:7 ratio. This provides a stricter evaluation of generalization to unseen proteins and ligands.

The results are shown in Figure~\ref{fig_crossdomain}. It is important to note that these experiments are restricted to interaction-free methods, as structural interaction information is unavailable in the simulated cross-domain and cold scenarios. Compared to the earlier in-domain results, the performance of all interaction-free PLA models drops significantly as shown in Figure~\ref{fig_crossdomain}a. This decline is primarily attributed to the reduced information overlap between the training and test sets, as well as the absence of sufficiently similar proteins and ligands to serve as references. Moreover, Figure~\ref{fig_crossdomain}b highlights a noticeable performance drop when moving from random split to cold pair split, emphasizing that prediction accuracy on unseen test data cannot solely rely on previously encountered protein or ligand features.

Despite the increased difficulty of cold-pair and cross-domain settings, KEPLA consistently outperforms DrugBAN across all evaluation scenarios. In particular, under the most challenging cross-domain setting, KEPLA achieves a lower RMSE and a higher Pearson's R than DrugBAN, with a 12.23\% reduction in RMSE and an 8.37\% improvement in R. Moreover, KEPLA maintains a higher Pearson's R under the cross-domain setting, suggesting better robustness to distribution shifts. These results demonstrate the generalization capability of KEPLA across different evaluation settings.

\subsection{Cross-Dataset Performance Comparison}

To further evaluate cross-dataset generalization, we construct an external test set based on the Astex Diverse Set~\cite{hartshorn2007diverse}. This dataset contains crystallized poses of 85 structurally diverse protein-ligand complexes and has been widely used for docking validation. Since standardized affinity labels are not available for all Astex complexes, only those with reliable affinity annotations from public records are retained. To prevent data leakage, PDB ID overlaps with this external test set are excluded from the original training and validation data.

As shown in Table~\ref{tab:astex_result}, KEPLA outperforms both interaction-free and interaction-based baselines on the Astex Diverse Set. This improvement suggests that the knowledge-enhanced representation learning strategy can provide complementary biochemical and semantic information beyond structure alone. By incorporating knowledge from both protein and ligand sides during representation learning, KEPLA learns more informative and transferable representations, which contribute to its improved performance in external evaluations and further support its cross-dataset generalization.

\begin{table}[t]
\centering
\setlength{\tabcolsep}{3pt}
\caption{Cross-dataset performance on Astex Diverse Set.}
\label{tab:astex_result}
\begin{tabular}{lcccc}
\toprule
Method & \multicolumn{4}{c}{Astex Diverse Set}\\
\cmidrule(lr){2-5}
 & RMSE $\downarrow$ & MAE $\downarrow$ & SD $\downarrow$ & R $\uparrow$ \\
\midrule
DeepDTA & 1.700 (0.033) & 1.359 (0.028) & 1.424 (0.028) & 0.637 (0.035)\\
GAT-GCN & 1.625 (0.024) & 1.299 (0.019) & 1.368 (0.043) & 0.660 (0.027)\\
DrugBAN & 1.436 (0.028) & 1.264 (0.019) & 1.347 (0.028) & 0.684 (0.028)\\
SIGN & 1.365 (0.023) & 1.036 (0.027) & 1.334 (0.042) & 0.691 (0.034)\\
GIANT & 1.294 (0.032) & \underline{1.011 (0.023)} & 1.302 (0.035) & 0.698 (0.018) \\
KSM & \underline{1.292 (0.027)} & 1.018 (0.019) & \underline{1.291 (0.034)} & \underline{0.715 (0.030)} \\
KEPLA & \pmb{1.265 (0.024)} & \pmb{0.993 (0.022)} & \pmb{1.255 (0.029)} & \pmb{0.734 (0.027)} \\
\bottomrule
\end{tabular}
\end{table}

\begin{table}[t]
\centering
\setlength{\tabcolsep}{2.2pt}
\caption{Ablation study in terms of RMSE and R on the PDBbind and CSAR-HiQ datasets under random split.}
\label{tab_ablation}\footnotesize
\begin{tabular}{lcccc}
\toprule
Variant & \multicolumn{2}{c}{PDBbind core set} & \multicolumn{2}{c}{CSAR-HiQ set} \\
\cmidrule(lr){2-3} \cmidrule(lr){4-5}
 & RMSE $\downarrow$ & R $\uparrow$ & RMSE $\downarrow$ & R $\uparrow$ \\
\midrule
w/o KG & 1.325 (0.022) & 0.790 (0.008) & 1.599 (0.030) & 0.763 (0.009)\\
Protein KG & 1.234 (0.023) & 0.816 (0.004) & 1.489 (0.013) & 0.794 (0.018)\\
Ligand KG & 1.288 (0.023) & 0.803 (0.009) & 1.561 (0.034) & 0.764 (0.016)\\
\midrule
Linear concat. & 1.291 (0.011) & 0.804 (0.002) & 1.559 (0.026) & 0.760 (0.011)\\
Protein attn. & 1.266 (0.049) & 0.815 (0.005) & 1.574 (0.026) & 0.767 (0.014)\\
Ligand attn. & 1.248 (0.025) & 0.816 (0.005) & 1.536 (0.019) & 0.777 (0.023)\\
\midrule
w/o protein enc. & 1.263 (0.016) & 0.813 (0.004) & 1.570 (0.028) & 0.778 (0.015)\\
w/o ligand enc. & 1.319 (0.020) & 0.797 (0.010) & 1.584 (0.029) & 0.774 (0.019)\\
Linear dec. & 1.408 (0.019) & 0.757 (0.006) & 1.592 (0.035) & 0.761 (0.017)\\
\midrule
KEPLA & \pmb{1.202 (0.017)} & \pmb{0.832 (0.005)} & \pmb{1.459 (0.019)} & \pmb{0.813 (0.012)} \\
\bottomrule
\end{tabular}
\end{table}

\subsection{Ablation Study}
We conduct ablation studies to investigate the contributions of the KG, cross attention module, and backbone components in KEPLA. To evaluate KG enhancement, we compare KEPLA with three variants: KEPLA using only the protein-side KG, KEPLA using only the ligand-side KG, and KEPLA without KG enhancement. As shown in Table~\ref{tab_ablation}, the performance drops after removing or partially using the KG, validating the effectiveness of KG enhancement. Notably, the protein-side KG contributes more significantly than the ligand-side counterpart. A likely explanation is that the protein KG contains more entities and richer relational information, offering more comprehensive context for representation learning.

To assess the cross attention mechanism, we evaluate three KEPLA variants that differ in how they compute joint representations between proteins and ligands: one-side protein attention, one-side ligand attention, and linear concatenation. In these variants, the cross attention module in KEPLA is replaced with either ligand-side or protein-side attention. The one-side attention variants adopt the neural attention mechanism proposed in~\cite{tsubaki2019compound}, where the joint representation is derived from the interaction between a ligand vector and a protein fragment matrix. The linear concatenation variant simply concatenates the protein and ligand vector representations following a max-pooling operation. As shown in Table~\ref{tab_ablation}, the results confirm the effectiveness of the cross attention mechanism, as it enables the model to capture fine-grained interactions crucial for accurate PLA prediction. Among the one-side attention variants, ligand-side attention yields better performance. This may be because ligand-side attention benefits from local 2D molecular representations.

For the backbone component ablation, \textit{w/o Protein Encoder} replaces the ESM encoder with three consecutive 1D convolutional layers following DrugBAN~\cite{bai2023interpretable}, \textit{w/o Ligand Encoder} replaces the GCN-based ligand encoder with a DNN-based ligand encoder, and \textit{Linear Decoder} replaces the MLP decoder with a single linear layer. As shown in Table~\ref{tab_ablation}, all variants underperform KEPLA on both datasets, confirming the contribution of these core components. The large drop of \textit{Linear Decoder} on the PDBbind core set indicates the importance of nonlinear prediction layers, while the degradation caused by replacing GCN with DNN shows the necessity of graph-based ligand representation.

\begin{figure*}[htbp]
\centering
\includegraphics[width=1\linewidth]{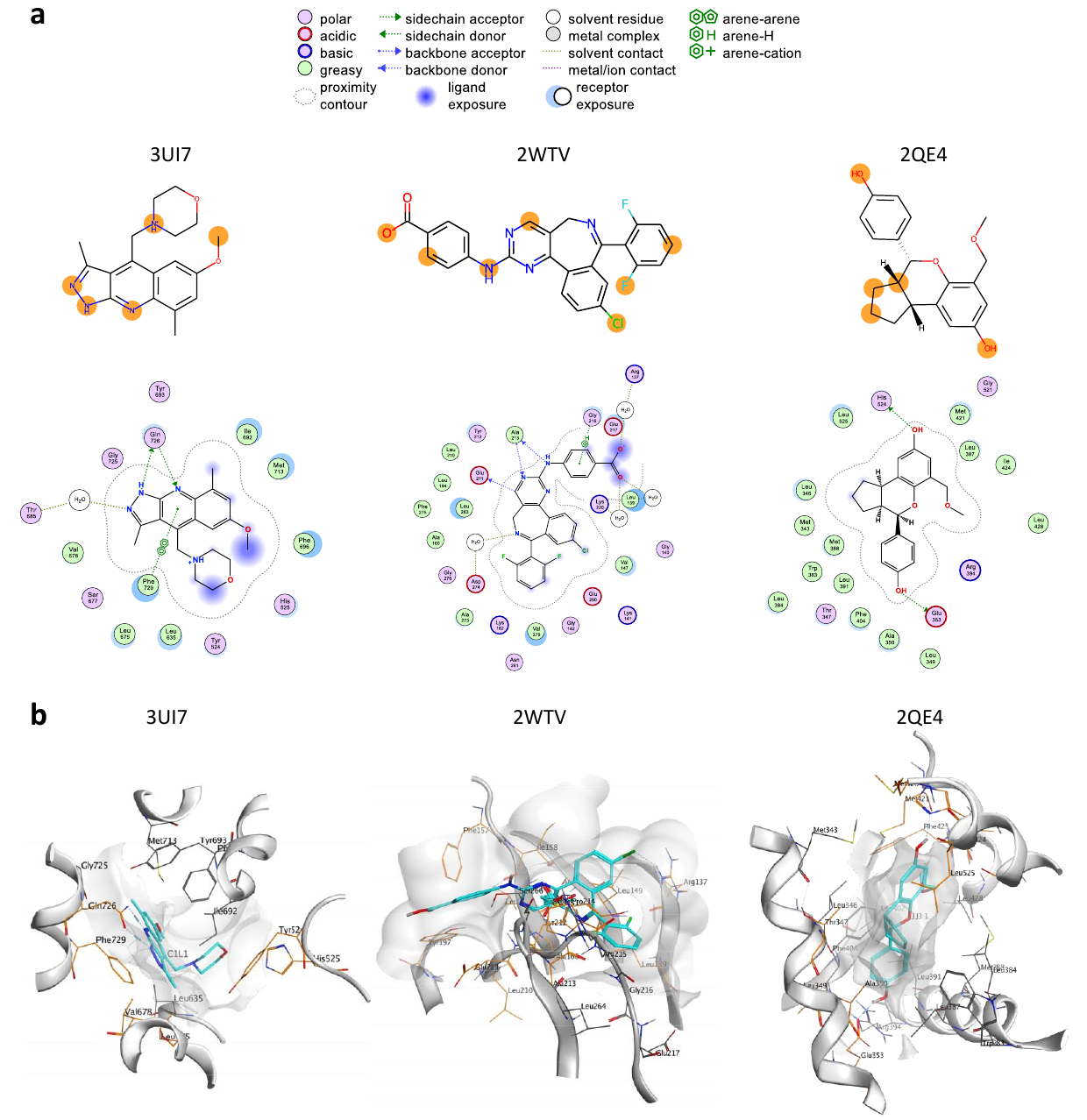}
\caption{\textbf{Visualization of ligands and binding pockets for structural-level interpretability study.} \textbf{(a)}  Interpretability of co-crystallized ligands. The upper section of each panel displays the 2D structures of ligands, with atoms highlighted in orange to indicate those predicted to contribute to protein binding. All 2D structures are visualized using RDKit. The lower section of each panel presents protein-ligand interaction maps derived from the corresponding crystal structures. \textbf{(b)}  Interpretability of binding pocket structures. The 3D representations show protein-ligand binding pockets, with correctly predicted binding-site residues highlighted in orange and the corresponding ligands shown in cyan. Remaining amino acid residues, secondary structure elements, and surface maps are shown in grey. All protein-ligand interaction maps and 3D visualizations of X-ray crystal structures are generated using the Molecular Operating Environment (MOE) software.}\label{fig_attn_expl}
\end{figure*}

\subsection{Structural-Level Interpretability with Cross Attention}
A further strength of KEPLA lies in its ability to provide interpretability at both structural and knowledge levels---an essential feature for drug design applications. Structural-level interpretability is achieved by leveraging cross attention weights to visualize the contribution of each substructure to the final prediction. To illustrate this, we examine three top-ranked predictions (PDB IDs: 3UI7, 2WTV, and 2QE4) corresponding to co-crystallized ligands from the Protein Data Bank (PDB)~\cite{berman2000protein}. 
Only X-ray structures with a resolution greater than $2.0\mathring{\text{A}}$ and targeting human proteins are considered~\cite{maveyraud2020protein}. 
Additionally, the selected complexes must exhibit binding affinities at the nanoscale and belong to the test set. Visualization results are presented in Figure~\ref{fig_attn_expl}a, alongside protein-ligand interaction maps derived from the corresponding X-ray structures. For each ligand, the top 20\% of atoms, ranked by cross attention weights, are highlighted in orange.

For 3UI7, KEPLA correctly identifies nitrogen atoms within the heterocyclic core as key hydrogen-bond donors and acceptors, matching experimental interactions with Gln729 and Thr685. The pyridine ring involved in an arene-arene contact with Phe729 is also partly emphasized. Notably, most of the top-ranked atoms correspond to verified interaction sites. For 2WTV, the model highlights major binding motifs, including the aniline amino group (hydrogen bond donor to Ala213), carboxylate (interacting with Tyr212), and nitrogen-containing heterocycles (hydrogen bond to Glu211). Despite the ligand’s higher complexity, KEPLA accurately captures most key interactions observed in the crystal structure. For 2QE4, KEPLA pinpoints hydroxyl groups forming hydrogen bonds with Glu353 and His524 and identifies several solvent-exposed carbons potentially involved in secondary interactions.

Protein-level interpretability is coarser, as attention is computed over residue fragments rather than single residues. Even so, the model successfully highlights key binding residues: Gln726 and Phe729 in 3UI7, Glu211 and Ala213 in 2WTV, and Glu353 and His524 in 2QE4, along with other pocket-forming residues shown in Figure~\ref{fig_attn_expl}b. Across complexes, nearly half of the experimentally verified binding-site residues are detected, suggesting fragment-level attention provides a solid basis for interpretability. Future incorporation of residue-level attention and local 3D structural cues may further improve resolution and explanatory power.


\begin{figure*}[!htbp]
\centering
\includegraphics[width=1\linewidth]{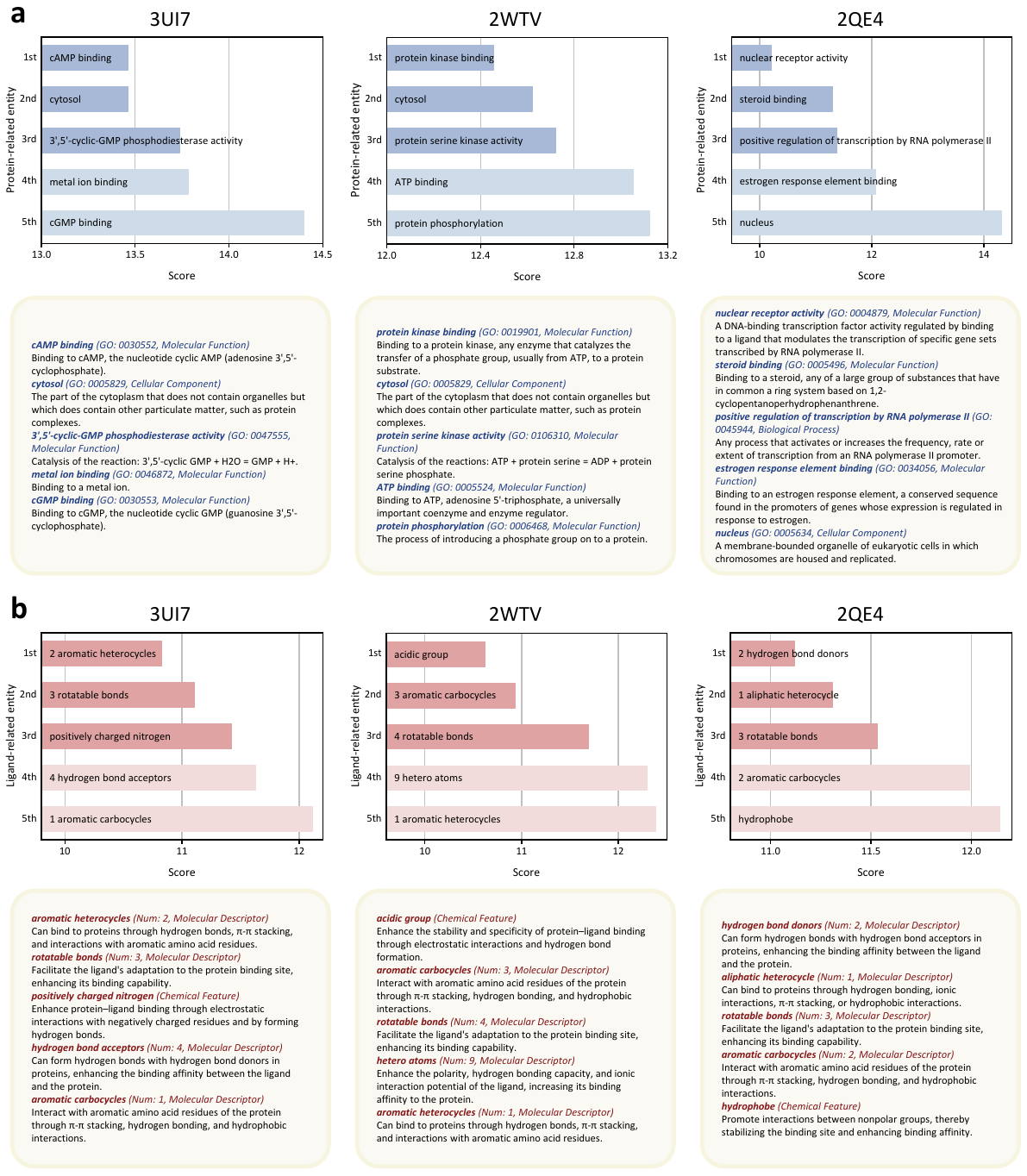}
\caption{\textbf{Biochemical knowledge of proteins and ligands for knowledge-level interpretability study.}~\textbf{(a)} Interpretability of protein biological knowledge. The upper part of each panel displays the GO entities most closely associated with the protein, ranked in ascending order according to their triplet scores from the KG embedding model. The lower part presents the types and detailed definitions of these entities, annotated with their corresponding GO IDs.~\textbf{(b)} Interpretability of ligand chemical knowledge. The upper part of each panel shows the LP entities most closely related to the ligand, also ranked in ascending order by their triplet scores in the KG embedding model. The lower part provides the types and specific functions of these entities, with molecular descriptors annotated by their corresponding numbers.}\label{fig_kg_vis}
\end{figure*}

\subsection{Knowledge-Level Interpretability with Knowledge Graph}

We further examine whether external knowledge can enhance the interpretability of protein-ligand binding. Specifically, we identify entities most closely related to the target protein or ligand by applying the KG scoring function to all associated triplets. Lower scores indicate stronger semantic relevance. As illustrated in Figure~\ref{fig_kg_vis}, the top five nearest GO or LP entities and their definitions are used to explain the binding mechanism. 

For 3UI7 (pyrazoloquinoline bound to human PDE10A), the top GO entities most closely associated with the protein include \textit{cAMP binding}, \textit{cytosol}, \textit{3',5'-cyclic-GMP phosphodiesterase activity}, \textit{metal ion binding}, and \textit{cGMP binding}. 
These entities are consistent with the known phosphodiesterase function of PDE10A, including cyclic nucleotide hydrolysis and metal-ion-dependent catalytic activity. This also agrees with the binding of pyrazoloquinoline in the catalytic pocket of PDE10A, where it interferes with substrate access and interacts with the metal-ion-associated catalytic environment.
On the ligand side, the top associated entities for pyrazoloquinoline are \textit{two aromatic heterocycles}, \textit{three rotatable bonds}, \textit{positively charged nitrogen}, \textit{four hydrogen bond acceptors}, and \textit{one aromatic carbocycle}. 
These descriptors suggest potential aromatic interactions, hydrogen bonding, and moderate conformational flexibility, which are also consistent with the attention-based structural explanation in Figure~\ref{fig_attn_expl}a.


For 2WTV (MLN8054 bound to human Aurora kinase A), the top GO entities most closely associated with the protein include \textit{protein kinase binding}, \textit{cytosol}, \textit{protein serine kinase activity}, \textit{ATP binding}, and \textit{protein phosphorylation}. These annotations are consistent with the kinase function of Aurora kinase A. The interaction between MLN8054 and Aurora kinase A aligns with these annotations. This also agrees with MLN8054 acting as an ATP-competitive Aurora kinase inhibitor. On the ligand side, the LP entities most closely associated with MLN8054 include \textit{acidic group}, \textit{three aromatic carbocycles}, \textit{four rotatable bonds}, \textit{nine heteroatoms}, and \textit{one aromatic heterocycle}. They facilitate hydrogen bonding, metal coordination, and conformational flexibility, consistent with high binding affinity.

For 2QE4 (a benzopyran agonist complexed with the human estrogen receptor), top-ranked GO entities such as \textit{nuclear receptor activity}, \textit{steroid binding}, and \textit{estrogen response element binding} reflect the receptor’s steroid hormone-related function. Ligand-side LP entities, including \textit{two hydrogen bond donors}, \textit{two aromatic carbocycles}, and \textit{hydrophobe}, indicate possible hydrogen bonding and hydrophobic interactions that stabilize the complex.

These examples show that KEPLA can retrieve biologically and chemically meaningful KG entities, providing interpretable explanations for PLA predictions.

\subsection{Train-Test Redundancy Analysis}
To further investigate whether the performance is affected by train-test redundancy, we conduct a redundancy analysis at both protein and ligand levels. For each test complex, we compute the minimum PSC-based cosine distance between its protein and all proteins in the training set, and the minimum ECFP4-based Jaccard distance between its ligand and all ligands in the training set. Following the clustering-based split protocol, proteins and ligands are regarded as redundant when their distances are below 0.001 and 0.5, respectively.

\begin{figure}[t]
    \centering

    \begin{minipage}[t]{0.492\linewidth}
        \centering
        \includegraphics[width=\linewidth]{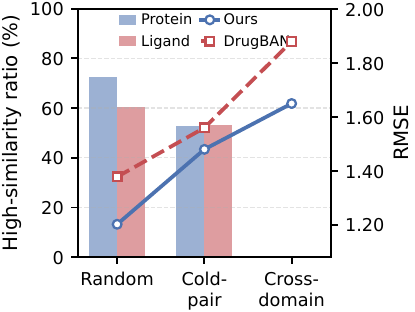}
    \end{minipage}
    \hfill
    \begin{minipage}[t]{0.492\linewidth}
        \centering
        \includegraphics[width=\linewidth]{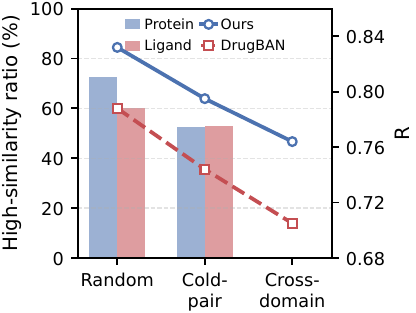}
    \end{minipage}

    \caption{Train-test redundancy analysis. The bars show protein- and ligand-level redundancy ratios, and the lines show the corresponding RMSE and R results.}
    \label{fig:redundancy_analysis}
\end{figure}

Figure~\ref{fig:redundancy_analysis} summarizes the redundancy ratios and the corresponding predictive performance under different evaluation settings. The bars show the proportions of redundant test proteins and ligands, while the lines report the RMSE and R values of KEPLA and the strongest interaction-free baseline DrugBAN. The random split shows high protein- and ligand-level redundancy, indicating that many test samples have similar counterparts in the training set. The cold-pair split reduces such redundancy but still retains a certain proportion of similar proteins and ligands. In contrast, the clustering-based cross-domain split removes both protein- and ligand-level redundancy under the predefined thresholds, resulting in a more challenging evaluation setting.

As the redundancy decreases from the random split to the cross-domain split, both methods show increased RMSE and decreased R, which confirms that reducing train-test similarity makes prediction more difficult. Nevertheless, KEPLA consistently achieves lower RMSE and higher R than the strongest baseline across all settings. These results suggest that KEPLA does not merely rely on highly similar training samples, but maintains stronger generalization ability under reduced train-test redundancy and cross-domain distribution shift.

\subsection{Generalization Ability Comparison}
\begin{figure}[t]
\centering
\includegraphics[width=\linewidth]{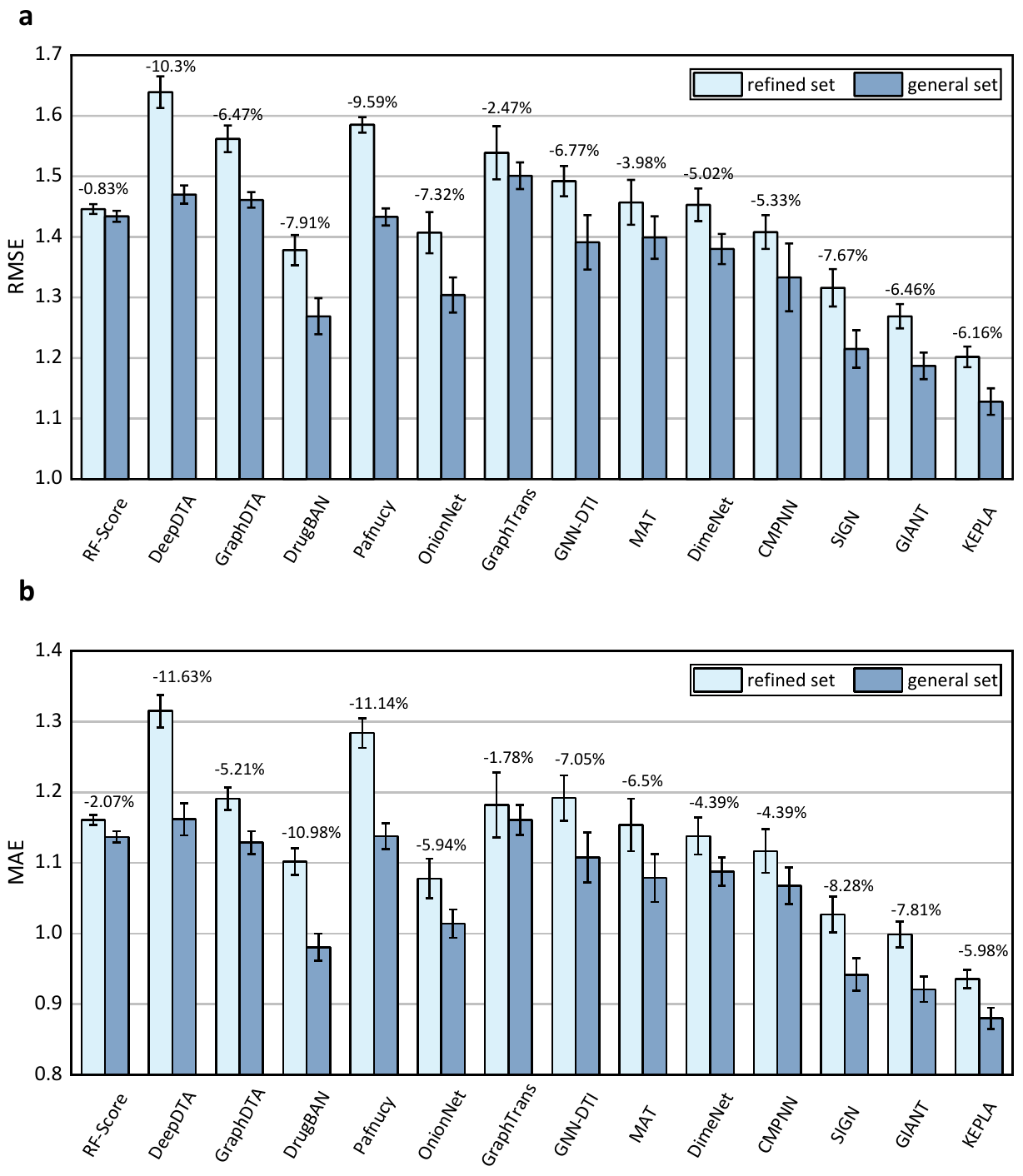}
\caption{\textbf{Performance improvements on PDBbind benchmark when trained on the \textit{general set}.} \textbf{(a)} RMSE reduction on PDBbind core set. \textbf{(b)} MAE reduction on PDBbind core set.}\label{fig_general}
\end{figure}

As PLA data continues to grow, the proportion of high-quality data in the \textit{refined set} remains relatively limited. Consequently, a model’s ability to effectively utilize a larger volume of lower-quality data becomes a critical measure of its generalization capability---an essential aspect of performance evaluation. To assess this, we conduct additional generalization experiments on the PDBbind \textit{general set}, which contains all 13,283 protein-ligand complexes. As shown in Figure~\ref{fig_general}, KEPLA consistently achieves the lowest prediction errors under both training settings compared to mainstream baseline models. Notably, when trained on the \textit{general set}, KEPLA reduces RMSE and MAE by approximately 6.16\% and 5.98\%, respectively, further widening the performance gap with baseline models. These results highlight KEPLA’s strong generalization ability to large-scale, lower-quality data.

\section{Conclusion}
In this work, we present KEPLA, an end-to-end knowledge-enhanced deep learning framework for PLA prediction, designed to address the performance bottlenecks and limited interpretability of existing methods. Specifically, we construct a biochemical KG for proteins and ligands, respectively. The encoded representations of proteins and ligands are integrated into the KG embedding objective to capture the rich semantic information within KG. Subsequently, the model employs cross attention mechanisms to perform fine-grained aggregation of the knowledge-enhanced representations, which are then fed into a decoder to predict binding affinity. We conduct extensive experiments to evaluate the effectiveness of KEPLA in addressing the performance and interpretability challenges.

In future work, we plan to incorporate 3D structural information into our PLA prediction models, leveraging recent advancements like AlphaFold~\cite{abramson2024accurate}, which has predicted over 2 billion protein structures. This integration has the potential to further improve model performance and interpretability. Additionally, expanding our knowledge graph with more comprehensive domain-specific and cross-disciplinary knowledge could enhance prediction accuracy. We also aim to adapt the KEPLA framework to other molecular interaction prediction tasks, such as drug-drug and protein-protein interactions~\cite{su2022biomedical,ji2012survey}, demonstrating its versatility and broader applicability.

\section*{Acknowledgments}
This work was supported in part by the Institute of Digital Medicine, City University of Hong Kong; the CityUHK Strategic Seed Fund; Frontiers Medical Center, Tianfu Jincheng Laboratory; and the National Natural Science Foundation of China under Grant 62572282.



 
\bibliography{reference}

@article{li2021dgl,
  title={Dgl-lifesci: An open-source toolkit for deep learning on graphs in life science},
  author={Li, Mufei and Zhou, Jinjing and Hu, Jiajing and Fan, Wenxuan and Zhang, Yangkang and Gu, Yaxin and Karypis, George},
  journal={ACS Omega},
  volume={6},
  number={41},
  pages={27233--27238},
  year={2021},
  publisher={ACS Publications}
}

@article{bordes2013translating,
  title={Translating embeddings for modeling multi-relational data},
  author={Bordes, Antoine and Usunier, Nicolas and Garcia-Dur{\'a}n, Alberto and Weston, Jason and Yakhnenko, Oksana},
  journal={International Conference on Neural Information Processing Systems},
  pages={2787--2795},
  year={2013}
}

@article{DBLP:journals/corr/KingmaB14,
  author       = {Diederik P. Kingma and
                  Jimmy Ba},
  title        = {Adam: {A} Method for Stochastic Optimization},
  journal    = {International Conference on Learning Representations},
  year         = {2015},
}

@article{nguyen2021graphdta,
  title={GraphDTA: predicting drug--target binding affinity with graph neural networks},
  author={Nguyen, Thin and Le, Hang and Quinn, Thomas P and Nguyen, Tri and Le, Thuc Duy and Venkatesh, Svetha},
  journal={Bioinformatics},
  volume={37},
  number={8},
  pages={1140--1147},
  year={2021},
  publisher={Oxford University Press}
}

@article{ozturk2018deepdta,
  title={DeepDTA: deep drug--target binding affinity prediction},
  author={{\"O}zt{\"u}rk, Hakime and {\"O}zg{\"u}r, Arzucan and Ozkirimli, Elif},
  journal={Bioinformatics},
  volume={34},
  number={17},
  pages={i821--i829},
  year={2018},
  publisher={Oxford University Press}
}

@article{ballester2010machine,
  title={A machine learning approach to predicting protein--ligand binding affinity with applications to molecular docking},
  author={Ballester, Pedro J and Mitchell, John BO},
  journal={Bioinformatics},
  volume={26},
  number={9},
  pages={1169--1175},
  year={2010},
  publisher={Oxford University Press}
}

@article{stepniewska2018development,
  title={Development and evaluation of a deep learning model for protein--ligand binding affinity prediction},
  author={Stepniewska-Dziubinska, Marta M and Zielenkiewicz, Piotr and Siedlecki, Pawel},
  journal={Bioinformatics},
  volume={34},
  number={21},
  pages={3666--3674},
  year={2018},
  publisher={Oxford University Press}
}

@article{zheng2019onionnet,
  title={Onionnet: a multiple-layer intermolecular-contact-based convolutional neural network for protein--ligand binding affinity prediction},
  author={Zheng, Liangzhen and Fan, Jingrong and Mu, Yuguang},
  journal={ACS Omega},
  volume={4},
  number={14},
  pages={15956--15965},
  year={2019},
  publisher={ACS Publications}
}

@article{lim2019predicting,
  title={Predicting drug--target interaction using a novel graph neural network with 3D structure-embedded graph representation},
  author={Lim, Jaechang and Ryu, Seongok and Park, Kyubyong and Choe, Yo Joong and Ham, Jiyeon and Kim, Woo Youn},
  journal={Journal of Chemical Information and Modeling},
  volume={59},
  number={9},
  pages={3981--3988},
  year={2019},
  publisher={ACS Publications}
}

@article{weininger1988smiles,
  title={SMILES, a chemical language and information system. 1. Introduction to methodology and encoding rules},
  author={Weininger, David},
  journal={Journal of Chemical Information and Computer Sciences},
  volume={28},
  number={1},
  pages={31--36},
  year={1988},
  publisher={ACS Publications}
}

@article{tsubaki2019compound,
  title={Compound--protein interaction prediction with end-to-end learning of neural networks for graphs and sequences},
  author={Tsubaki, Masashi and Tomii, Kentaro and Sese, Jun},
  journal={Bioinformatics},
  volume={35},
  number={2},
  pages={309--318},
  year={2019},
  publisher={Oxford University Press}
}

@article{volkov2022frustration,
  title={On the frustration to predict binding affinities from protein--ligand structures with deep neural networks},
  author={Volkov, Mikhail and Turk, Joseph-Andr{\'e} and Drizard, Nicolas and Martin, Nicolas and Hoffmann, Brice and Gaston-Math{\'e}, Yann and Rognan, Didier},
  journal={Journal of Medicinal Chemistry},
  volume={65},
  number={11},
  pages={7946--7958},
  year={2022},
  publisher={ACS Publications}
}

@article{luo2017network,
  title={A network integration approach for drug-target interaction prediction and computational drug repositioning from heterogeneous information},
  author={Luo, Yunan and Zhao, Xinbin and Zhou, Jingtian and Yang, Jinglin and Zhang, Yanqing and Kuang, Wenhua and Peng, Jian and Chen, Ligong and Zeng, Jianyang},
  journal={Nature Communications},
  volume={8},
  number={1},
  pages={573},
  year={2017},
  publisher={Nature Publishing Group UK London}
}

@article{lee2019deepconv,
  title={DeepConv-DTI: Prediction of drug-target interactions via deep learning with convolution on protein sequences},
  author={Lee, Ingoo and Keum, Jongsoo and Nam, Hojung},
  journal={PLoS Computational Biology},
  volume={15},
  number={6},
  pages={1--21},
  year={2019},
  publisher={Public Library of Science San Francisco, CA USA}
}

@article{yang2022mgraphdta,
  title={MGraphDTA: deep multiscale graph neural network for explainable drug--target binding affinity prediction},
  author={Yang, Ziduo and Zhong, Weihe and Zhao, Lu and Chen, Calvin Yu-Chian},
  journal={Chemical Science},
  volume={13},
  number={3},
  pages={816--833},
  year={2022},
  publisher={Royal Society of Chemistry}
}

@article{vaswani2017attention,
  title={Attention is all you need},
  author={Vaswani, Ashish and Shazeer, Noam and Parmar, Niki and Uszkoreit, Jakob and Jones, Llion and Gomez, Aidan N and Kaiser, {\L}ukasz and Polosukhin, Illia},
  journal={International Conference on Neural Information Processing Systems},
  pages = {1--11},
  year={2017}
}

@article{chen2020transformercpi,
  title={TransformerCPI: improving compound--protein interaction prediction by sequence-based deep learning with self-attention mechanism and label reversal experiments},
  author={Chen, Lifan and Tan, Xiaoqin and Wang, Dingyan and Zhong, Feisheng and Liu, Xiaohong and Yang, Tianbiao and Luo, Xiaomin and Chen, Kaixian and Jiang, Hualiang and Zheng, Mingyue},
  journal={Bioinformatics},
  volume={36},
  number={16},
  pages={4406--4414},
  year={2020},
  publisher={Oxford University Press}
}

@article{huang2021moltrans,
  title={MolTrans: molecular interaction transformer for drug--target interaction prediction},
  author={Huang, Kexin and Xiao, Cao and Glass, Lucas M and Sun, Jimeng},
  journal={Bioinformatics},
  volume={37},
  number={6},
  pages={830--836},
  year={2021},
  publisher={Oxford University Press}
}

@article{ashburner2000gene,
  title={Gene ontology: tool for the unification of biology},
  author={Ashburner, Michael and Ball, Catherine A and Blake, Judith A and Botstein, David and Butler, Heather and Cherry, J Michael and Davis, Allan P and Dolinski, Kara and Dwight, Selina S and Eppig, Janan T and others},
  journal={Nature Genetics},
  volume={25},
  number={1},
  pages={25--29},
  year={2000},
  publisher={Nature Publishing Group}
}

@article{wang2017knowledge,
  title={Knowledge graph embedding: A survey of approaches and applications},
  author={Wang, Quan and Mao, Zhendong and Wang, Bin and Guo, Li},
  journal={IEEE Transactions on Knowledge and Data Engineering},
  volume={29},
  number={12},
  pages={2724--2743},
  year={2017},
  publisher={IEEE}
}

@article{ye2021unified,
  title={A unified drug--target interaction prediction framework based on knowledge graph and recommendation system},
  author={Ye, Qing and Hsieh, Chang-Yu and Yang, Ziyi and Kang, Yu and Chen, Jiming and Cao, Dongsheng and He, Shibo and Hou, Tingjun},
  journal={Nature Communications},
  volume={12},
  number={1},
  pages={6775},
  year={2021},
  publisher={Nature Publishing Group UK London}
}

@article{hou2019cross,
  title={Cross attention network for few-shot classification},
  author={Hou, Ruibing and Chang, Hong and Ma, Bingpeng and Shan, Shiguang and Chen, Xilin},
  journal={International Conference on Neural Information Processing Systems},
  pages={4003--4014},
  year={2019}
}

@article{sadybekov2023computational,
  title={Computational approaches streamlining drug discovery},
  author={Sadybekov, Anastasiia V and Katritch, Vsevolod},
  journal={Nature},
  volume={616},
  number={7958},
  pages={673--685},
  year={2023},
  publisher={Nature Publishing Group UK London}
}

@article{li2025knowledge,
  title={Knowledge-enhanced and structure-enhanced representation learning for protein-ligand binding affinity prediction},
  author={Li, Mei and Cao, Ye and Liu, Xiaoguang and Ji, Hua},
  journal={Pattern Recognition},
  volume={166},
  pages={111701},
  year={2025},
  publisher={Elsevier}
}

@article{mastropietro2023learning,
  title={Learning characteristics of graph neural networks predicting protein--ligand affinities},
  author={Mastropietro, Andrea and Pasculli, Giuseppe and Bajorath, J{\"u}rgen},
  journal={Nature Machine Intelligence},
  volume={5},
  number={12},
  pages={1427--1436},
  year={2023},
  publisher={Nature Publishing Group UK London}
}

@article{lai2024interformer,
  title={Interformer: an interaction-aware model for protein-ligand docking and affinity prediction},
  author={Lai, Houtim and Wang, Longyue and Qian, Ruiyuan and Huang, Junhong and Zhou, Peng and Ye, Geyan and Wu, Fandi and Wu, Fang and Zeng, Xiangxiang and Liu, Wei},
  journal={Nature Communications},
  volume={15},
  number={1},
  pages={10223},
  year={2024},
  publisher={Nature Publishing Group UK London}
}

@article{berman2000protein,
  title={The protein data bank},
  author={Berman, Helen M and Westbrook, John and Feng, Zukang and Gilliland, Gary and Bhat, Talapady N and Weissig, Helge and Shindyalov, Ilya N and Bourne, Philip E},
  journal={Nucleic Acids Research},
  volume={28},
  number={1},
  pages={235--242},
  year={2000},
  publisher={Oxford University Press}
}

@article{maveyraud2020protein,
  title={Protein X-ray crystallography and drug discovery},
  author={Maveyraud, Laurent and Mourey, Lionel},
  journal={Molecules},
  volume={25},
  number={5},
  pages={1030},
  year={2020},
  publisher={MDPI}
}

@article{abramson2024accurate,
  title={Accurate structure prediction of biomolecular interactions with AlphaFold 3},
  author={Abramson, Josh and Adler, Jonas and Dunger, Jack and Evans, Richard and Green, Tim and Pritzel, Alexander and Ronneberger, Olaf and Willmore, Lindsay and Ballard, Andrew J and Bambrick, Joshua and others},
  journal={Nature},
  volume={630},
  number={8016},
  pages={493--500},
  year={2024},
  publisher={Nature Publishing Group UK London}
}

@article{lin2023evolutionary,
  title={Evolutionary-scale prediction of atomic-level protein structure with a language model},
  author={Lin, Zeming and Akin, Halil and Rao, Roshan and Hie, Brian and Zhu, Zhongkai and Lu, Wenting and Smetanin, Nikita and Verkuil, Robert and Kabeli, Ori and Shmueli, Yaniv and others},
  journal={Science},
  volume={379},
  number={6637},
  pages={1123--1130},
  year={2023},
  publisher={American Association for the Advancement of Science}
}

@article{sunrotate,
  title={RotatE: Knowledge Graph Embedding by Relational Rotation in Complex Space},
  author={Sun, Zhiqing and Deng, Zhi-Hong and Nie, Jian-Yun and Tang, Jian},
  journal={International Conference on Learning Representations}
}

@article{bai2023interpretable,
  title={Interpretable bilinear attention network with domain adaptation improves drug--target prediction},
  author={Bai, Peizhen and Miljkovi{\'c}, Filip and John, Bino and Lu, Haiping},
  journal={Nature Machine Intelligence},
  volume={5},
  number={2},
  pages={126--136},
  year={2023},
  publisher={Nature Publishing Group UK London}
}

@article{li2024giant,
  title={Giant: Protein-ligand binding affinity prediction via geometry-aware interactive graph neural network},
  author={Li, Shuangli and Zhou, Jingbo and Xu, Tong and Huang, Liang and Wang, Fan and Xiong, Haoyi and Huang, Weili and Dou, Dejing and Xiong, Hui},
  journal={IEEE Transactions on Knowledge and Data Engineering},
  volume={36},
  number={5},
  pages={1991--2008},
  year={2024},
  publisher={IEEE}
}

@article{li2021structure,
  title={Structure-aware interactive graph neural networks for the prediction of protein-ligand binding affinity},
  author={Li, Shuangli and Zhou, Jingbo and Xu, Tong and Huang, Liang and Wang, Fan and Xiong, Haoyi and Huang, Weili and Dou, Dejing and Xiong, Hui},
  journal={ACM SIGKDD Conference on Knowledge Discovery and Data Mining},
  pages={975--985},
  year={2021}
}

@article{danel2020spatial,
  title={Spatial graph convolutional networks},
  author={Danel, Tomasz and Spurek, Przemys{\l}aw and Tabor, Jacek and {\'S}mieja, Marek and Struski, {\L}ukasz and S{\l}owik, Agnieszka and Maziarka, {\L}ukasz},
  journal={International Conference on Neural Information Processing},
  pages={668--675},
  year={2020},
  organization={Springer}
}

@article{wu2021representing,
  title={Representing long-range context for graph neural networks with global attention},
  author={Wu, Zhanghao and Jain, Paras and Wright, Matthew and Mirhoseini, Azalia and Gonzalez, Joseph E and Stoica, Ion},
  journal={International Conference on Neural Information Processing Systems},
  volume={34},
  pages={13266--13279},
  year={2021}
}

@article{gasteiger_dimenet_2020,
title = {Directional Message Passing for Molecular Graphs},
author = {Gasteiger, Johannes and Gro{\ss}, Janek and G{\"u}nnemann, Stephan},
journal={International Conference on Learning Representations},
year = {2020} }

@article{song2020communicative,
  title={Communicative representation learning on attributed molecular graphs},
  author={Song, Ying and Zheng, Shuangjia and Niu, Zhangming and Fu, Zhang-Hua and Lu, Yutong and Yang, Yuedong},
  journal={International Joint Conference on Artificial Intelligence},
  pages={2831--2838},
  year={2020},
}

@article{maziarka2020molecule,
  title={Molecule attention transformer},
  author={Maziarka, {\L}ukasz and Danel, Tomasz and Mucha, S{\l}awomir and Rataj, Krzysztof and Tabor, Jacek and Jastrz{\k{e}}bski, Stanis{\l}aw},
  journal={arXiv preprint arXiv:2002.08264},
  year={2020}
}

@article{yang2019analyzing,
  title={Analyzing learned molecular representations for property prediction},
  author={Yang, Kevin and Swanson, Kyle and Jin, Wengong and Coley, Connor and Eiden, Philipp and Gao, Hua and Guzman-Perez, Angel and Hopper, Timothy and Kelley, Brian and Mathea, Miriam and others},
  journal={Journal of Chemical Information and Modeling},
  volume={59},
  number={8},
  pages={3370--3388},
  year={2019},
  publisher={ACS Publications}
}

@article{liu2021non,
  title={Non-local graph neural networks},
  author={Liu, Meng and Wang, Zhengyang and Ji, Shuiwang},
  journal={IEEE Transactions on Pattern Analysis and Machine Intelligence},
  volume={44},
  number={12},
  pages={10270--10276},
  year={2021},
  publisher={IEEE}
}

@article{liu2015pdb,
  title={PDB-wide collection of binding data: current status of the PDBbind database},
  author={Liu, Zhihai and Li, Yan and Han, Li and Li, Jie and Liu, Jie and Zhao, Zhixiong and Nie, Wei and Liu, Yuchen and Wang, Renxiao},
  journal={Bioinformatics},
  volume={31},
  number={3},
  pages={405--412},
  year={2015},
  publisher={Oxford University Press}
}

@article{smith2011csar,
  title={CSAR benchmark exercise of 2010: combined evaluation across all submitted scoring functions},
  author={Smith, Richard D and Dunbar Jr, James B and Ung, Peter Man-Un and Esposito, Emilio X and Yang, Chao-Yie and Wang, Shaomeng and Carlson, Heather A},
  journal={Journal of Chemical Information and Modeling},
  volume={51},
  number={9},
  pages={2115--2131},
  year={2011},
  publisher={ACS Publications}
}

@article{li2022dyscore,
  title={DyScore: a boosting scoring method with dynamic properties for identifying true binders and nonbinders in structure-based drug discovery},
  author={Li, Yanjun and Zhou, Daohong and Zheng, Guangrong and Li, Xiaolin and Wu, Dapeng and Yuan, Yaxia},
  journal={Journal of Chemical Information and Modeling},
  volume={62},
  number={22},
  pages={5550--5567},
  year={2022},
  publisher={ACS Publications}
}

@article{rezaei2020deep,
  title={Deep learning in drug design: protein-ligand binding affinity prediction},
  author={Rezaei, Mohammad A and Li, Yanjun and Wu, Dapeng and Li, Xiaolin and Li, Chenglong},
  journal={IEEE/ACM Transactions on Computational Biology and Bioinformatics},
  volume={19},
  number={1},
  pages={407--417},
  year={2020},
  publisher={IEEE}
}

@article{jiang2020drug,
  title={Drug--target affinity prediction using graph neural network and contact maps},
  author={Jiang, Mingjian and Li, Zhen and Zhang, Shugang and Wang, Shuang and Wang, Xiaofeng and Yuan, Qing and Wei, Zhiqiang},
  journal={RSC advances},
  volume={10},
  number={35},
  pages={20701--20712},
  year={2020},
  publisher={Royal Society of Chemistry}
}

@article{ma2022kg,
  title={Kg-mtl: knowledge graph enhanced multi-task learning for molecular interaction},
  author={Ma, Tengfei and Lin, Xuan and Song, Bosheng and Yu, Philip S and Zeng, Xiangxiang},
  journal={IEEE Transactions on Knowledge and Data Engineering},
  volume={35},
  number={7},
  pages={7068--7081},
  year={2022},
  publisher={IEEE}
}

@article{su2022biomedical,
  title={Biomedical knowledge graph embedding with capsule network for multi-label drug-drug interaction prediction},
  author={Su, Xiaorui and You, Zhuhong and Huang, Deshuang and Wang, Lei and Wong, Leon and Ji, Boya and Zhao, Bowei},
  journal={IEEE Transactions on Knowledge and Data Engineering},
  volume={35},
  number={6},
  pages={5640--5651},
  year={2022},
  publisher={IEEE}
}

@article{ji2012survey,
  title={Survey: Functional module detection from protein-protein interaction networks},
  author={Ji, Junzhong and Zhang, Aidong and Liu, Chunnian and Quan, Xiaomei and Liu, Zhijun},
  journal={IEEE Transactions on Knowledge and Data Engineering},
  volume={26},
  number={2},
  pages={261--277},
  year={2012},
  publisher={IEEE}
}

@article{hartshorn2007diverse,
  title={Diverse, high-quality test set for the validation of protein- ligand docking performance},
  author={Hartshorn, Michael J and Verdonk, Marcel L and Chessari, Gianni and Brewerton, Suzanne C and Mooij, Wijnand TM and Mortenson, Paul N and Murray, Christopher W},
  journal={Journal of medicinal chemistry},
  volume={50},
  number={4},
  pages={726--741},
  year={2007},
  publisher={ACS Publications}
}
%

\bibliographystyle{IEEEtran}

\vspace{-33pt}
\begin{IEEEbiography}[{\includegraphics[width=1in,height=1.25in,clip,keepaspectratio]{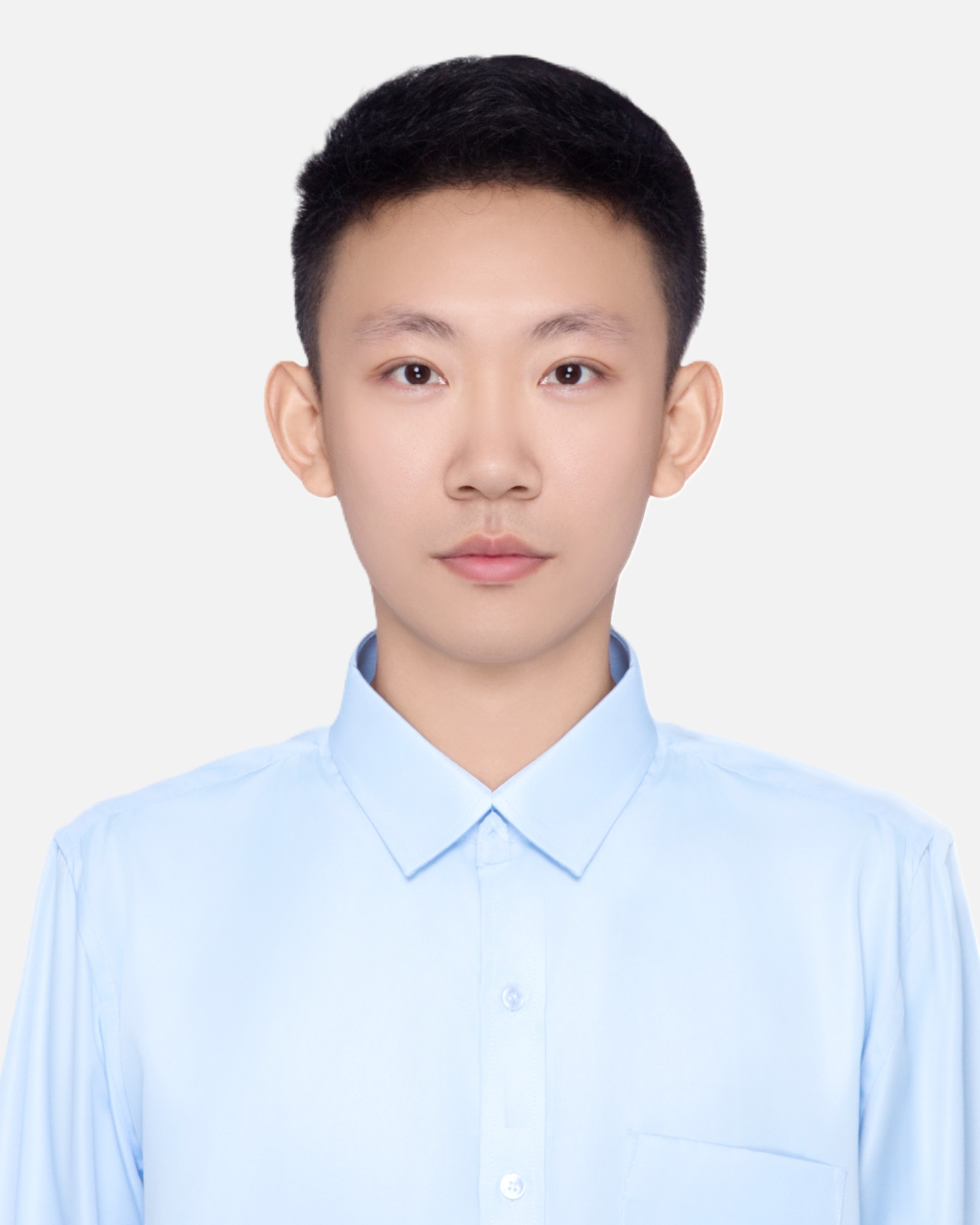}}]{Han Liu} received the B.E. and M.E. degrees in software engineering, and the Ph.D. degree in computer science and technology from Shandong University. He is currently a Postdoctoral Fellow in the Department of Computer Science at City University of Hong Kong. His research interests include AI for science, multimodal retrieval, and recommender systems.
\end{IEEEbiography}

\begin{IEEEbiography}[{\includegraphics[width=1in,height=1.25in,clip,keepaspectratio]{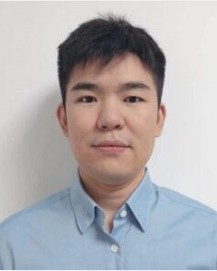}}]{Keyan Ding} received the Ph.D. degree from the City University of Hong Kong in 2021. He is currently a Hundred Talents Program Researcher at the ZJU-Hangzhou Global Scientific and Technological Innovation Center, Zhejiang University. He was selected for the Zhejiang Provincial Overseas Talent Program in 2024. His research interests include computer vision, natural language processing, and AI for Science. He has authored or coauthored more than 30 papers as the first or corresponding author in prestigious journals and conferences, such as Nature Machine Intelligence, Nature Computational Science, IEEE Transactions on Pattern Analysis and Machine Intelligence, ICLR, and ICML.

\end{IEEEbiography}

\begin{IEEEbiography}[{\includegraphics[width=1in,height=1.25in,clip,keepaspectratio]{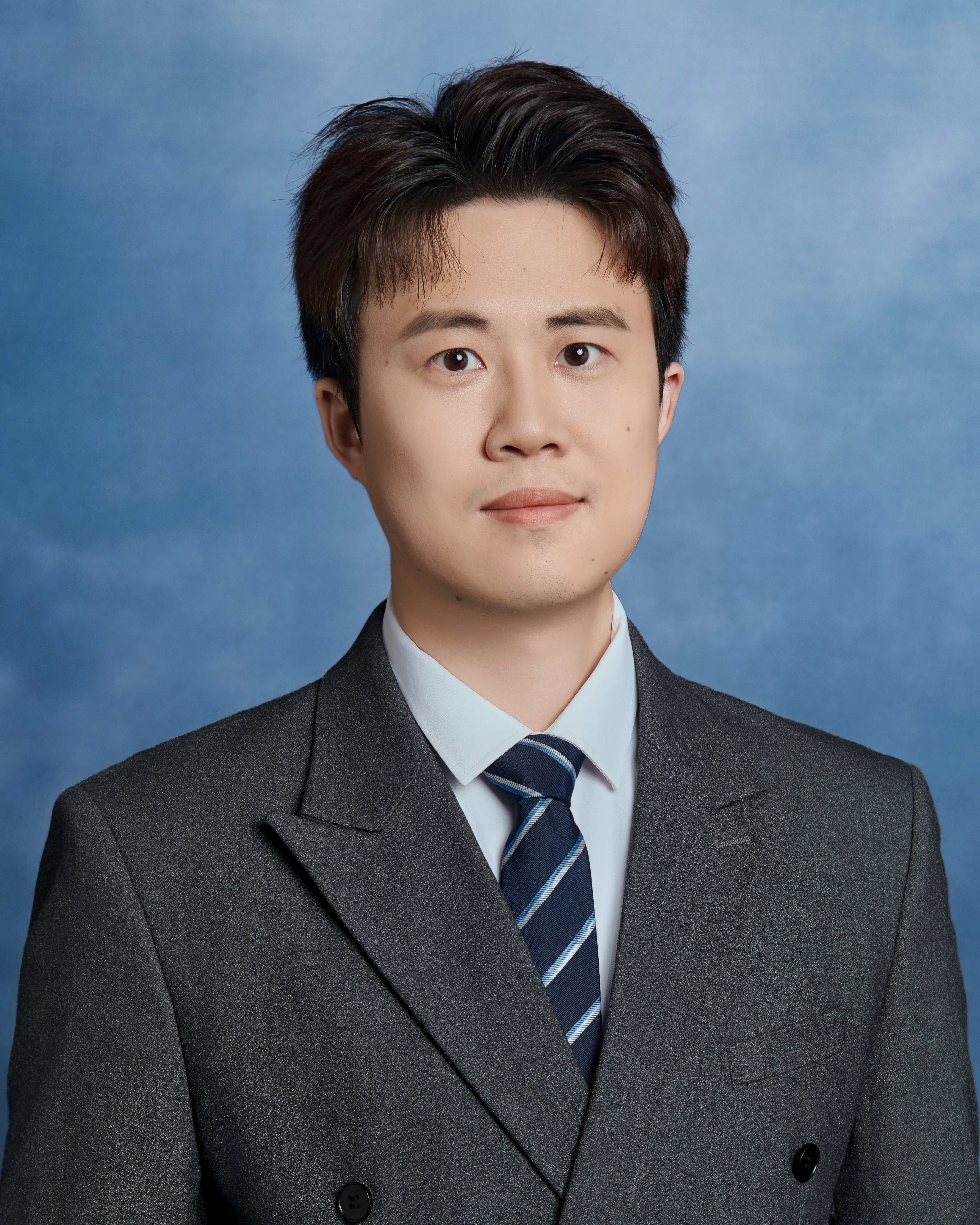}}]{Peilin Chen} received the B.E. degree in software engineering from Sun Yat-sen University (SYSU) in 2018 and his Ph.D. degree in computer science from the City University of Hong Kong (CityUHK) in 2023. He is currently a Postdoctoral Researcher in the Department of Computer Science at CityUHK. His research interests include visual data processing/enhancement and multimodal large language models.
\end{IEEEbiography}

\begin{IEEEbiography}[{\includegraphics[width=1in,height=1.25in,clip,keepaspectratio]{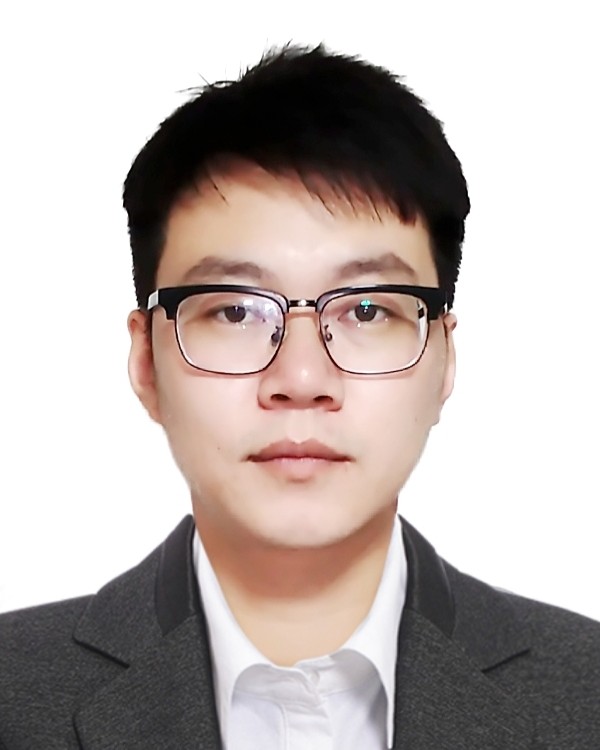}}]{Yinwei Wei} is currently a professor with Shandong University, China. He received his Ph.D. degree from Shandong University. His research interests include multimodal content analysis and information retrieval. Several works have been published in top forums, such as IEEE TPAMI, TIP, and TMM. Dr. Wei has served as the AC or PC member for several conferences, such as NeurIPS, ICML, and SIGIR, and the regular reviewer for TMM, TKDE, and TIP.
\end{IEEEbiography}

\begin{IEEEbiography}[{\includegraphics[width=1in,height=1.25in,clip,keepaspectratio]{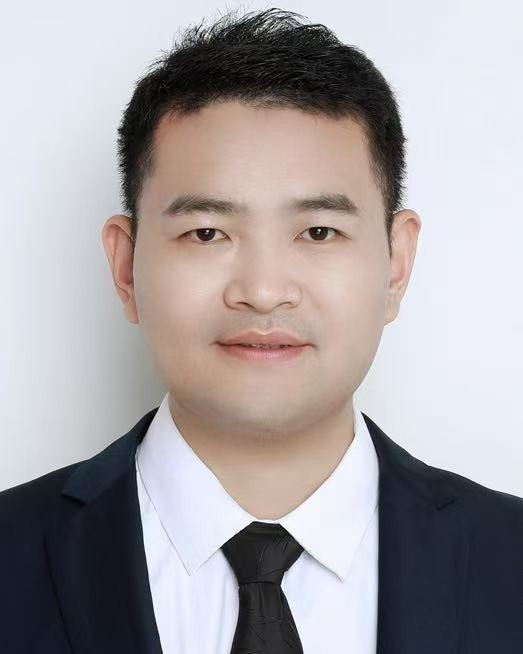}}]{Liqiang Nie} (Senior Member, IEEE) received his B.Eng. and Ph.D. degree from Xi'an Jiaotong University and National University of Singapore (NUS), respectively. His research interest is multimodal large language models, embodied intelligence, and information retrieval. Dr. Nie has co-/authored more than 200 papers and 5 books, with 40,477 Google Scholar citations. He is an AE of IEEE T-PAMI, IEEE T-KDE, IEEE T-MM, IEEE T-CSVT, ACM ToMM, and Information Science. Meanwhile, he is the pc chair of ICMR 2025, ICME 2025, and ACM MM 2027. He is a member of ICME steering committee. He has received many awards over the past five years, like ACM MM and SIGIR best paper honorable mention in 2019, the AI 2000 most influential scholars 2020, SIGMM rising star in 2020, MIT TR35 China 2020, DAMO Academy Young Fellow in 2020, SIGIR best student paper in 2021, the first price of the provincial science and technology progress award in 2021 and 2023 (rank 1), IEEE AI’s 10 to Watch in 2022, ACM MM Best paper award in 2022, national youth science and technology award in 2024, and several first place awards of ground challenges organized by CCF A conferences. Some of his research outputs have been integrated into the products of Alibaba, Kwai, and other listed companies. He is an IAPR Fellow.
\end{IEEEbiography}

\begin{IEEEbiography}[{\includegraphics[width=1in,height=1.25in,clip,keepaspectratio]{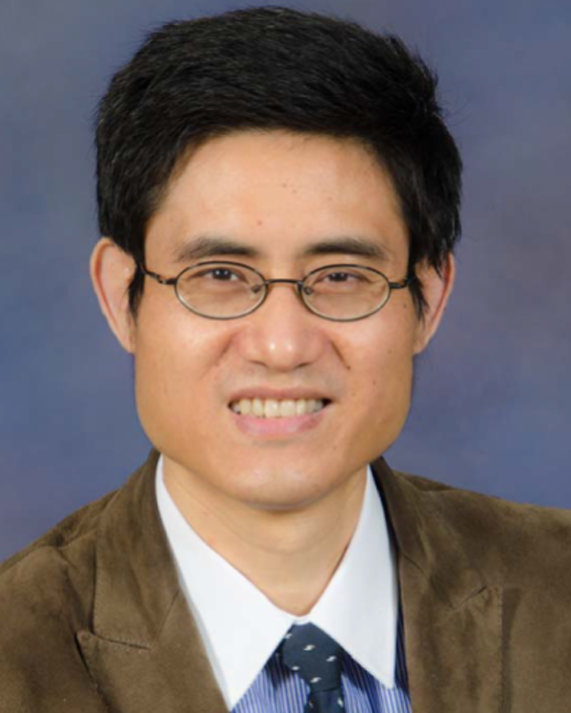}}]{Dapeng Wu} (Fellow, IEEE) received the B.E. degree in electrical engineering from the Huazhong University of Science and Technology, Wuhan, China, in 1990, the M.E. degree in electrical engineering from Beijing University of Posts and Telecommunications, Beijing, China, in 1997, and the Ph.D. degree in electrical and computer engineering from Carnegie Mellon University, Pittsburgh, PA, USA, in 2003. He was on the faculty of University of Florida, Gainesville, FL, USA, and served as the Director of the NSF Center for Big Learning, USA. He is currently the Yeung Kin Man Chair Professor of network science and a Chair Professor of data engineering with the Department of Computer Science, City University of Hong Kong. His research interests include artificial intelligence, network science, communications, signal processing, computer vision, and biomedical engineering. He received the NSF CAREER Award, multiple young investigator program awards, university professorship awards, and multiple IEEE journal and conference best paper awards. He has served as the TPC Chair for IEEE INFOCOM and IEEE ICC and a committee member for over 100 conferences. He has served as the Founding Editor-in-Chief and the Editor-in-Chief for several academic and IEEE journals, and also as an associate editor for several IEEE Transactions and magazines. He was elected as a Distinguished Lecturer by IEEE Vehicular Technology Society in 2016. He is an IEEE Fellow.
\end{IEEEbiography}

\begin{IEEEbiography}[{\includegraphics[width=1in,height=1.25in,clip,keepaspectratio]{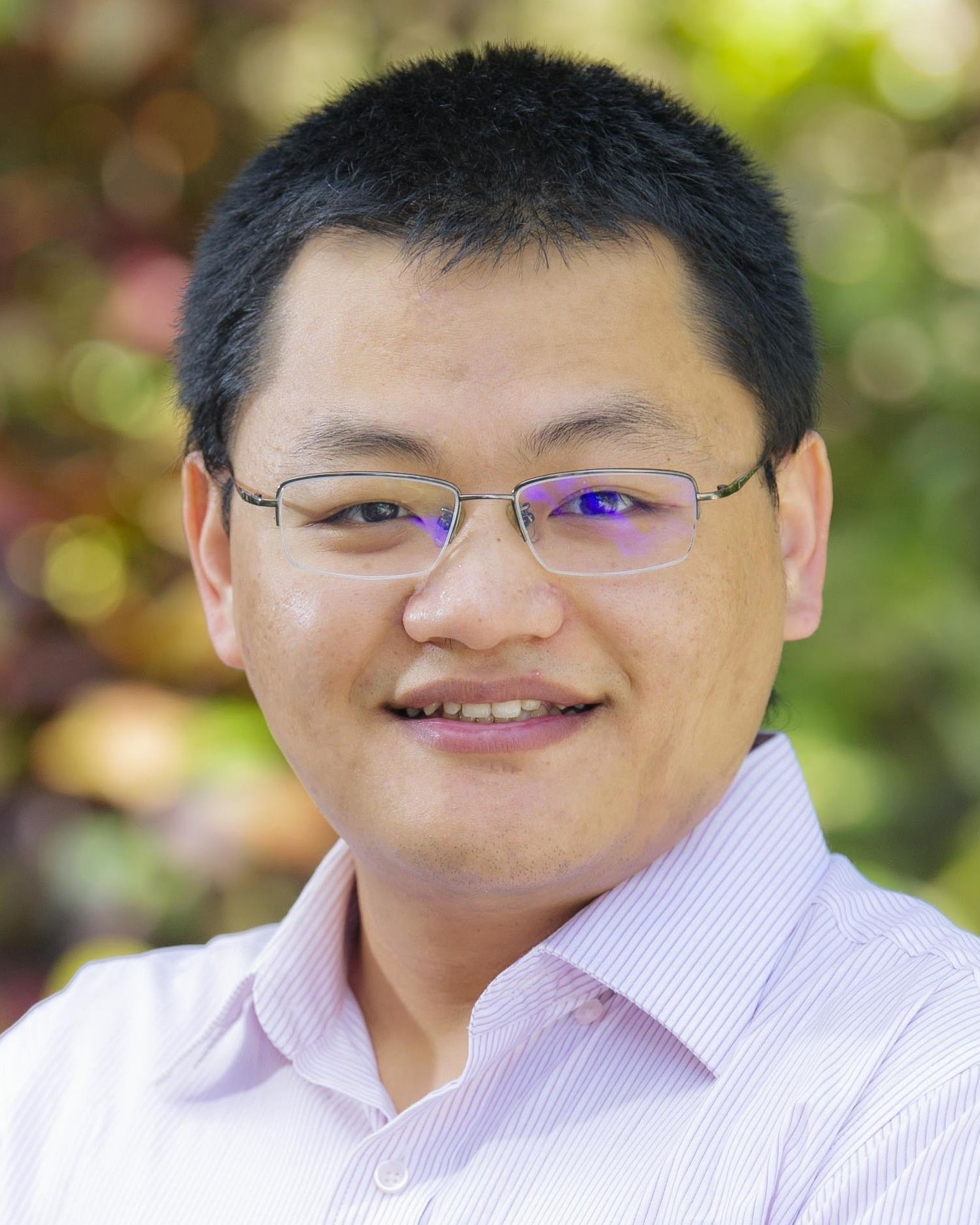}}]{Shiqi Wang} (Senior Member, IEEE) received the PhD degree in computer application technology from Peking University in 2014. He is currently a Professor with the Department of Computer Science, City University of Hong Kong, Hong Kong. He has proposed more than 70 technical proposals to ISO/MPEG, ITUT, and AVS standards. He authored or coauthored more than 300 refereed journal articles/conference papers, including more than 100 IEEE Transactions. His research interests include semantic and visual communication, AI generated content management, machine learning, information forensics and security, and image/video quality assessment. He received the Best Paper Award from IEEE VCIP 2019, ICME 2019, IEEE Multimedia 2018, and PCM 2017. His coauthored article received the Best Student Paper Award in the IEEE ICIP 2018. He was the TPC Chair of ICME 2024. He served or serves as an associate editor for IEEE Transactions on Circuits and Systems for Video Technology, IEEE Transactions on Multimedia, IEEE Transactions on Image Processing, and IEEE Transactions on Cybernetics.
\end{IEEEbiography}




\end{document}